\documentclass{article}

\usepackage{microtype}
\usepackage{graphicx}
\usepackage{subfigure}
\usepackage{caption}
\usepackage{booktabs} % for professional tables
\usepackage{hyperref}
\usepackage{math_command}

\usepackage[accepted]{icml2024}

\usepackage{amsmath}
\usepackage{amssymb}
\usepackage{mathtools}
\usepackage{amsthm}
\usepackage{multirow}

\usepackage[capitalize,noabbrev]{cleveref}

\theoremstyle{plain}
\newtheorem{theorem}{Theorem}[section]
\newtheorem{proposition}[theorem]{Proposition}

\theoremstyle{definition}

\newtheorem{remark}{Remark}[section]

\usepackage[textsize=tiny]{todonotes}

\icmltitlerunning{Adaptive Preference Scaling for RLHF}

\begin{document}
\onecolumn
%\twocolumn[
\icmltitle{Adaptive Preference Scaling for Reinforcement Learning with Human Feedback}



\icmlsetsymbol{equal}{*}

\begin{icmlauthorlist}
\icmlauthor{Ilgee Hong}{equal,gatech}
\icmlauthor{Zichong Li}{equal,gatech}
\icmlauthor{Alexander Bukharin}{gatech}
\icmlauthor{Yixiao Li}{gatech}
\icmlauthor{Haoming Jiang}{amazon}
\icmlauthor{Tianbao Yang}{tamu}
\icmlauthor{Tuo Zhao}{gatech}
\end{icmlauthorlist}

\icmlaffiliation{gatech}{Georgia Tech}
\icmlaffiliation{amazon}{Amazon}
\icmlaffiliation{tamu}{Texas A\&M University}

\icmlcorrespondingauthor{Ilgee Hong, Zichong Li, Tuo Zhao}{ihong39,zli911,tourzhao@gatech.edu}

\icmlkeywords{Machine Learning, ICML}

\vskip 0.3in



\printAffiliationsAndNotice{\icmlEqualContribution} % otherwise use the standard text.

% Learning from human preferences has become a crucial stage in developing AI systems to make them understand and adapt to human preferences.
\begin{abstract}
Reinforcement learning from human feedback (RLHF) is a prevalent approach to align AI systems with human values by learning rewards from human preference data. Due to various reasons, however, such data typically takes the form of rankings over pairs of trajectory segments, which fails to capture the varying strengths of preferences across different pairs. In this paper, we propose a novel adaptive preference loss, underpinned by distributionally robust optimization (DRO), designed to address this uncertainty in preference strength. By incorporating an adaptive scaling parameter into the loss for each pair, our method increases the flexibility of the reward function. Specifically, it assigns small scaling parameters to pairs with ambiguous preferences, leading to more comparable rewards, and large scaling parameters to those with clear preferences for more distinct rewards. Computationally, our proposed loss function is strictly convex and univariate with respect to each scaling parameter, enabling its efficient optimization through a simple second-order algorithm. Our method is versatile and can be readily adapted to various preference optimization frameworks, including direct preference optimization (DPO). Our experiments with robotic control and natural language generation with large language models (LLMs) show that our method not only improves policy performance but also aligns reward function selection more closely with policy optimization, simplifying the hyperparameter tuning process.
\end{abstract}

\section{Introduction}\label{sec1: intro}

In the field of artificial intelligence, aligning AI systems with human preferences has become increasingly crucial, particularly for applications involving complex data and models like large language models (LLMs) in natural language processing \citep{learn_summ_2020, instructGPT}. Reinforcement learning from human feedback (RLHF) has gained popularity for customizing AI systems \citep{christiano2017deep, HH_rlhf, hinge_2023}. RLHF involves learning a reward function from human preference data, then using a reinforcement learning algorithm to train a policy to optimize the learned reward model.

A key challenge in RLHF lies in the complexity of reward modeling, which primarily stems from the reliance on preference labels. Since preference labels only provide comparative rankings of trajectory segments without quantifying the scale of underlying preference strengths, previous methods have employed the Bradley-Terry (BT) model \citep{bradley1952rank} in conjunction with cross-entropy loss to learn the reward function from preference data \citep{christiano2017deep, learn_summ_2020}. This approach assumes that the logit of the preference distribution scales linearly with the reward difference across all sample pairs. However, such linear scaling is often insufficient to account for the variations in preference strength among different pairs, restricting the reward function's ability to capture a broader range of reward differences. This restrictive approach to reward modeling limits the flexibility of the learned reward function, hindering its capacity to produce the versatile rewards essential for the downstream policy optimization.

To overcome this shortcoming, we introduce a novel adaptive preference loss function inspired by distributionally robust optimization (DRO, \citet{duchi2021statistics}). Our approach incorporates an instance-specific scaling factor to change the scaling between the preference distribution and the reward difference to be non-linear. These factors are learned during training and enable the model to accommodate varying uncertainties of preference strength, thereby enhancing the flexibility of the reward. For pairs showing strong preference (i.e., low preference uncertainty), our method learns a large scaling factor, which enables the model to learn a larger reward difference. In contrast, for pairs showing ambiguous preferences (i.e., high preference uncertainty), our method assigns a smaller scaling factor, enabling the model to learn a smaller reward difference. The additional computational overhead of involving this scaling factor into training is negligible, as the proposed loss function is strictly convex and univariate with respect to each scaling parameter. Therefore, it can be easily optimized by a simple second-order algorithm within a few iterations. %\textcolor{blue}{We}

Our experiments on robotic control tasks \cite{mujoco} demonstrate that our method can learn a more flexible reward function, resulting in an improved policy. Surprisingly, we also discover that our method better aligns the learned reward function with downstream policy optimization. Specifically, when tuning hyperparameters for reward modeling, the simplest approach is to select the reward model according to preference prediction accuracy. However, the selected reward function (with the highest accuracy) often yields a downstream policy with poor performance. To address this misalignment, we usually have to jointly tune the parameters across both stages according to downstream policy performance, resulting in significant computational burden and tuning effort. Our proposed method can mitigate this misalignment: When using our adaptive loss, we can select the reward model based on preference prediction accuracy alone and yield a reasonably well-performing policy. This allows separate tuning of the two stages, easing tuning overhead. To our knowledge, the challenge of this misalignment issue is almost untouched in the RLHF literature, and we are the first to propose a principal approach to mitigate this issue.

Moreover, our method is generalizable and can be applied to other preference optimization algorithms. For instance, we implement it with direct preference optimization (DPO, \citet{rafailov2023direct}) and evaluate its effectiveness on natural language generation tasks using Llama-2 7B \citep{touvron2023llama}. Our results demonstrate that integrating adaptive preference scaling into DPO boosts policy performance, while preserving the benefits of alignment. Alignment is especially critical in this setting, where we employ proprietary models like Claude 2 \citep{claude} as judges for policy selection, which demands substantial costs for using the APIs. In the case without access to LLM assessment, we must select policy based solely on preference accuracy, under which our approach substantially outperforms other baselines.

We summarize our main contributions as follows: \textbf{I}. We propose an adaptive preference loss function for RLHF, inspired by distributionally robust optimization and further extend it to DPO. By incorporating instance-specific scaling factors, our method enhances the flexibility of the reward model; \textbf{II}. Our method not only enhances the policy performance, but also better aligns the learned reward function with policy optimization, significantly easing the tuning efforts; \textbf{III}. We evaluate the effectiveness of our method on robotic control and natural language generation tasks. Our experimental results demonstrate that our adaptive preference loss can substantially improve the learned policy on both tasks and achieve better alignment between the reward model and policy optimization.

\section{Related Works}\label{sec2: rw}

\textbf{Loss Functions for Reward Learning.}
Prior work on this topic is very limited. For example, \citet{rewardcollapse} propose using different loss functions for strong and ambiguous preference data in natural language generation tasks. They apply heavy-tailed loss functions for open-ended questions, where preference ambiguity is desirable, and light-tailed loss functions for close-ended questions requiring clear-cut rewards. However, their approach requires knowing the question type a priori, necessitating extra labeling effort, and may fail for complex questions containing both open and closed aspects. \citet{hinge_2023} propose using a hinge loss, which results in zero gradient when the learned reward difference exceeds a margin of 1. This limits the ability to learn very large differences in rewards. \citet{ipo_2022} develop $\Psi$ Preference Optimization with Identity Mapping (IPO), which modifies DPO with a loss function matching the scaling of KL-divergence between the learned policy and the initial policy to avoid overfitting due to weak regularization. In contrast to prior work, our method is more broadly applicable to complex preference learning tasks without needing additional labeling or sacrificing the ability to learn arbitrarily large reward differences.

\textbf{Adaptive Temperature Scaling (ATS).} Temperature scaling (TS) aims to adjust the entropy of probabilistic models by rescaling their logit outputs before the softmax function is applied. This simple method not only enables confidence calibration \cite{guo2017calibration}, but also plays a vital role in various machine learning methods, including knowledge distillation \cite{hinton2015distilling}, reinforcement learning \cite{ma2017softmax}, and contrastive learning \cite{wang2021understanding}. Building on TS, adaptive temperature scaling (ATS) enhances flexibility by using instance-specific scalars. Most ATS method trains an additional network for predicting the temperature parameter, which is further integrated into the softmax operator to calibrate the prediction probabilities \cite{wang2020contextual,ding2021local,balanya2022adaptive,joy2023sample}. %,li2023curriculum}. 

In contrast to the aforementioned ATS methods, the proposed adaptive preference scaling (APS) is not designed for classical confidence calibration, but is crafted specifically to enhance the training process of reward function in RLHF. Consequently, the interpretations of scaling factors in ATS and APS are \textit{opposite}. In ATS, a larger scaling parameter is applied to data with higher uncertainty (e.g., data that the classifier is likely to misclassify), which reduces the magnitude of the corresponding logit. Conversely, in APS, a larger scaling factor is assigned to data with clearer preferences, resulting in a larger logit. This distinction clarifies why the scaling parameter in our approach does not correspond to the concept of ``temperature" from statistical physics. Additionally, we propose a principled framework for learning scaling parameter based on DRO, which avoids the complexities of designing specific temperature networks and does not rely on heuristically designed loss functions.

\textbf{Distributionally Robust Optimization (DRO).}
DRO is a technique that trains machine learning models to be robust against uncertainty in the data distribution. Specifically, DRO finds a solution that performs well under the worst-case distribution within a specified uncertainty set around the empirical data distribution \cite{ben2013robust,bertsimas2018data,kuhn2019wasserstein,sagawa2019distributionally,duchi2021statistics}. DRO has been applied in various AI/ML domains to improve generalization when the test distribution differs from the training distribution \cite{oren2019distributionally, gokhale2021semantically,michel2021modeling,broscheit2022distributionally,wen2022distributionally,qi2022stochastic}. Our framework is motivated by \citet{qi2022stochastic}, which tackles KL-constrained DRO problem. However, our approach differs in two significant ways. First, instead of using a single KL constraint for the entire training dataset, we apply a separate KL constraint to each individual training data. Second, since each training data involves just two distributional variables, we can use a deterministic method to optimize these efficiently. Note that while our proposed method is inspired by DRO, it serves a distinct purpose: improving reward learning in RLHF, which is \textit{orthogonal} to distributional robustness.

\section{Method}\label{sec3: methods}

In this section, we first outline the problem setup, derive the loss function with adaptive preference scaling, and then provide theoretical motivation for our proposed loss. At last, we present an optimization algorithm, extend the approach to direct preference optimization, and introduce a variant of our proposed loss that incorporates quadratic regularization.

\subsection{Problem Setup}
We consider a reward-free Markov decision process $\cM=(\cS, \cA, p, \gamma)$ with state $s\in\cS$, action $a\in\cA$, state transition function $p$, and discount factor $\gamma$. The ground truth reward function $r: \cS\times\cA\rightarrow \RR$ is assumed to be unknown, but only human preferences over pairs of trajectory segments are observed. A trajectory segment is a sequence of consecutive state and action~pairs $\traj=\{(s_m,a_m),(s_{m+1},a_{m+1}),\dots,(s_{k-1},a_{k-1})\}\in (\cS\times\cA)^{k-m}$. We denote $\traj_1\succ \traj_2$ to indicate that the human preferred trajectory segment $\traj_1$ over the trajectory segment $\traj_2$ and denote the preferred one with a subscript $w$ and the dispreferred one with a subscript $l$ (i.e., $z_w$ and $z_l$). Here, we are given a human preference dataset of trajectory segments $\cD_{\mathrm{pref}}=(\traj_{w,i}, \traj_{l,i})_{i=1}^N$. Our goal is to find a reward $\hat{r}(s,a)$, which is well-aligned with human preferences. Once we learn the reward, we then find a policy $\pi\in\Delta_{\cA}^{\cS}$ such that it maximizes the expected sum of discounted rewards,
\begin{align*}%\label{eq:policy}
\max_{\pi}\mathbb{E}_{z\sim\cD_{\pi}}\big[\hat{r}(z)\big], 
\end{align*}
where $\hat{r}(z)=\sum_{(s_t,a_t)\in\traj} \gamma^{t}\hat{r}(s_t,a_t)$ and $\cD_{\pi}$ denotes the stationary distribution of the state-action pair induced by $\pi$.

\subsection{Reward Learning with Adaptive Preference Scaling}

% For notational simplicity, throughout the rest of the paper, we denote $$r(z)=\sum_{(s_t,a_t)\in\traj} \gamma^{t}r(s_t,a_t).$$
We now focus on the reward learning phase in RLHF, a crucial stage for capturing human preferences across various trajectory segments. The standard reward learning procedure assumes that the reward function determines a preference distribution, also known as the Bradley-Terry (BT) model \cite{bradley1952rank}, 
\begin{align}\label{eq:BT}
p_{r}(\traj_w\succ \traj_l) = \sigmoid(r(\traj_w)-r(\traj_l)),
\end{align}
where $\sigmoid$ denotes the sigmoid function.
%Under the Bradley-Terry assumption~\cite{bradley1952rank}, the standard framework of RLHF models a preference distribution using  $r_{\bphi}$
The reward function is then learned by minimizing the expectation of negative log-likelihood of $r$ over the preference data \cite{christiano2017deep}:
\begin{align}\label{eq:rm}
\min_{r}\;\cL_{\mathrm{pref}}(r) = -\mathbb{E}_{(\traj_w,\traj_l)\sim\cD_{\mathrm{pref}}}\big[\log p_r(\traj_w\succ \traj_l)\big].
\end{align}
As can be seen from \eqref{eq:BT}, the BT model essentially assumes that the logit of the preference distribution $\sigmoid^{-1}(p_{r}(\traj_w\succ \traj_l))$ scales linearly with the reward difference, regardless of the specific pair of samples. Such linear scaling, however, does not necessarily fit the downstream policy learning well. For some applications, it may lead to a reward function that is not flexible enough to differentiate a pair of segments, which are supposed to have significantly different rewards.
To address this challenge, we propose an adaptive preference loss based on KL-constrained distributionally robust optimization formulation \cite{qi2022stochastic}, which can implicitly change the scaling between the logit and the reward difference to be non-linear. Specifically, given a pair of trajectory segments
$(\traj_1,\traj_2)$, we denote $d_{r}(\traj_1,\traj_2)=\mathbf{1}(\traj_1\succ\traj_2)\cdot (r(\traj_2)-r(\traj_1))$ and $p=(p_1,p_2)$. We define the following instance-level loss:
\begin{align}\label{eq:dro-rm1}
\ell_{r}(z_1,z_2):=\underset{p\in\Delta_2}{\max}\;p_1d_{r}(\traj_1,\traj_2)+p_2d_{r}(\traj_2,\traj_1)-\tau_0\mathrm{KL}(p,1/2)\quad\quad\mathrm{s.t.}\;\;\; \mathrm{KL}(p,1/2)\le\rho_{0},
\end{align}
where $\Delta_2=\{p\in\RR^2:p_1+p_2=1, 0\le p_1,p_2\le 1\}$, $1/2$ is denoted for the uniform distribution, and $\rho_0,\tau_0>0$ are shared prespecified parameters across all instances. $\mathrm{KL}(\cdot,\cdot)$ denotes the KL divergence. Note that without the KL-constraint, \eqref{eq:dro-rm1} is reduced to the cross-entropy loss with $\tau_0=1$. Unlike general KL-constrained DRO formulation, which considers a distribution $p$ over all training samples, the distributional variable $p$ in \eqref{eq:dro-rm1} is specifically associated with binary preference comparisons for each training sample (i.e., each pair of trajectory segments).

We then convert~\eqref{eq:dro-rm1} into an equivalent minimax formulation based on the Lagrangian duality,
\begin{align*}
%&\underset{p\in\Delta_2}{\max}\;\underset{\lambda\ge 0}{\min}\;p_1d_{r}(\traj_1,\traj_2)+p_2d_{r}(\traj_2,\traj_1)\\&\quad\quad\quad\quad\quad\quad\quad -\lambda(\mathrm{KL}(p,1/2)-\rho_{0})-\tau_0\mathrm{KL}(p,1/2)\\
%&\Leftrightarrow\underset{\lambda\ge 0}{\min}\;\underset{p\in\Delta_2}{\max}\;p_1d_{r}(\traj_1,\traj_2)+p_2d_{r}(\traj_2,\traj_1)\\&\quad\quad\quad\quad\quad\quad\quad\quad\quad\quad -(\lambda+\tau_0)(\mathrm{KL}(p,1/2)-\rho_{0}).
%&\Leftrightarrow
\underset{\lambda\ge 0}{\min}\;\underset{p\in\Delta_2}{\max}\;p_1d_{r}(\traj_1,\traj_2)+p_2d_{r}(\traj_2,\traj_1)-(\lambda+\tau_0)(\mathrm{KL}(p,1/2)-\rho_{0}),
\end{align*}
where  $\lambda$ is the Lagrange multiplier.
By defining $\tau$ as $\tau = \lambda + \tau_0$ and applying the optimality condition for $p$, we have
\begin{align}\label{eq:drpf}
\underset{\tau\ge\tau_0}{\min}\;-\tau\log p_{r,\tau}(\traj_w\succ \traj_l)+(\rho_0-\log 2)\tau,
\end{align}
where
\begin{align}\label{eq:temp-BT}
p_{r,\tau}(\traj_{w}\succ \traj_{l}) = \sigmoid\bigg(\dfrac{r(\traj_{w})-r(\traj_{l})}{\tau}\bigg).
\end{align}
%we do not use classifier but reward + \tau does not involved in downstream task + principled
%desiging or learning temperature network is difficult
%Discussion: prob is fixed (ours)
%problem unbounded (ill-defined) we have regularization 
%quadratic change the value of tau_min

% Note that \eqref{eq:temp-BT} has the scaling parameter $\tau$ is different from the BT model in \eqref{eq:BT} as the logit is no longer linear in the reward difference and $\tau$ is not necessarily constant. Additionally, we have a scaling parameter $\tau$, which depends on each sample. 
We refer to Appendix \ref{appen:der1} for the full derivation. Note that the preference scaling factor $\tau$ in \eqref{eq:drpf} and \eqref{eq:temp-BT} serves as the Lagrange multiplier of \eqref{eq:dro-rm1}. This scaling parameter $\tau$ is used specifically for training the reward function $r$, rather than calibrating the preference distribution $p_{r,\tau}(\traj_{w}\succ \traj_{l})$. The scaler $\tau$ is used exclusively during the reward learning phase and is no longer needed in subsequent policy optimization, where the reward function $r$ alone is used.

% Note that the scaling parameter $\tau$ acts as the Lagrange multiplier of \eqref{eq:dro-rm1}. We apply Equation \eqref{eq:temp-BT} specifically for training the reward function $r$, not for calibrating the preference distribution of \eqref{eq:temp-BT}. The scaling parameter $\tau$ is only used during reward learning and is no longer needed in downstream policy optimization, where we only use the reward function $r$.
% % where an additional scaling parameter $\tau$ is added.
% Note that in RLHF, we do not use the preference classifier \eqref{eq:temp-BT} in downstream policy optimization but only use the reward function $r$ to predict the rewards for the trajectory segments sampled from the policy. Once the reward function is learned, the scaling parameter $\tau$ is no longer needed. Essentially, our scaler $\tau$ acts as the Lagrange multiplier, primarily aiming to improve the learning of the reward function, instead of being used for calibrating the preference distribution in \eqref{eq:temp-BT}.

Moreover, the scaling parameter $\tau$ is defined to be an instance-specific parameter corresponding to the pair of trajectory segments $(\traj_{w}, \traj_{l})$. Therefore, when applying our adaptive loss to reward learning, for each pair $(\traj_{w,i}, \traj_{l,i})$, we need to define a corresponding scaling parameter denoted by $\tau_i$. The overall loss function over the training set $\cD_{\mathrm{pref}}$ is as follows: 
% Moreover, the scaling parameter $\tau$ is defined to be an instance-specific parameter corresponding the pair of trajectory segments $(\traj_{w}, \traj_{l})$. Therefore, when applying our adaptive loss to reward learning, for each pair $(\traj_{w,i}, \traj_{l,i})$, we need to define a corresponding scaling parameter denoted by $\tau_i$. The overall loss function over the training set $\cD_{\mathrm{pref}}$ is as follows: 
% \begin{align}\label{eq:adap-rlhf}
% &\underset{r,T\in\Omega^N}{\min}\cL(r,T)\nonumber\\
% &\quad\quad:=\frac{1}{N}\sum_{i=1}^N\big(-\tau_i\log p_{r,\tau_i}(\traj_{w,i}\succ \traj_{l,i}) +\rho\tau_i\big),
% \end{align}
\begin{align}\label{eq:adap-rlhf}
\underset{r,\tau_1,...,\tau_N\in\Omega}{\min}\frac{1}{N}\sum_{i=1}^N\ell_i(r,\tau_i):=\frac{1}{N}\sum_{i=1}^N\big(-\tau_i\log p_{r,\tau_i}(\traj_{w,i}\succ \traj_{l,i}) +\rho\tau_i\big),
\end{align}
where $T=(\tau_1,\dots,\tau_N)$, $\Omega=\{\tau:\tau_0\le\tau\le\tau_{\max}\}$ with $\tau_{\max}$ as another prespecified parameter, and $\rho=\rho_0-\log 2>-\log 2$. Here, we also involve an upper bound $\tau_{\max}>0$ in \eqref{eq:adap-rlhf}, and we will explain why it is needed in the next subsection.
 %$\tau_i$'s, \eqref{eq:adap-rlhf} allows for greater flexibility in learning the reward function $r$.

% Under the assumption $|d_{r}(\traj_1,\traj_2)|$ is bounded, we have $\tau\le\tau_{\max}$ for some $\tau_{\max}>0$~\citep{qi2022stochastic}. Thus, we use $\Omega=\{\tau:\tau_0\le\tau\le\tau_{\max}\}$ for our constraint set. 

\subsection{Theoretical Insights} \label{sec3.3:theo}
We next provide some theoretical insights on why the scaling parameter $\tau$ can help gain adaptivity by a proposition. For simplicity, we only consider a pair of trajectory segments. 
\begin{proposition}\label{prop:1}
    Assume we have a pair of trajectory segments $z_1, z_2$, and the preference distribution $p(z_1\succ z_2)=p^*\in(0,1)$, i.e., the probability, that $z_1$ is preferred over $z_2$, is $p^*$. 
    Consider the problem of minimizing the expectation of our adaptive loss function over the preference distribution:
\begin{align} \label{eq:exp_adaptive_loss}
\min_{r,\tau \in \Omega}-\tau p^*\log \big(\sigma\big((r(z_1)-r(z_2))/\tau\big)\big)-\tau (1-p^*)\log \big(\sigma\big((r(z_2)-r(z_1))/\tau\big)\big)+\rho\tau.
\end{align}
Then the minimizer $\tau^*$ and $r^*$ of the expected loss satisfy
\begin{align*}
&\tau^* = 
\begin{cases}
    \tau_0 & \text{if } -p^*\log (p^*)-
(1-p^*)\log (1-p^*)+\rho>0, \\
    \tau_{\mathrm{max}} & \text{if } -p^*\log (p^*)-
(1-p^*)\log (1-p^*)+\rho<0,
\end{cases}\\
&r^*(z_1)-r^*(z_2)=\tau^* \sigma^{-1}(p^*).
\end{align*}
Here, $\sigma^{-1}$ is the inverse of sigmoid function.
%\vspace{-0.1in}
\end{proposition}

Note that the expected loss \eqref{eq:exp_adaptive_loss} is only for easing theoretical analysis, as $p^*$ is not accessible in practice. From Proposition~\ref{prop:1}, we can see that when $p^*$ is close enough to $0.5$, (i.e., the uncertainty of preference is large), the corresponding optimal $\tau^*$ is at the lower bound $\tau_{0}$.
The resulting optimal reward difference is $\tau_{0} \sigma^{-1}(p^*)$, which is smaller 
than the counterpart obtained by the cross-entropy loss when $\tau_0< 1$. Conversely, when $p^*$ is close to $0$ or $1$, (i.e., the uncertainty of preference is small), the resulting optimal $\tau^*$ is at the upper bound $\tau_\mathrm{max}$. Here, we introduce the upper bound $\tau_{\rm max}$ to ensure that the optimal $\tau^*$ is bounded. The resulting reward difference in this case is $\tau_{\max} \sigma^{-1}(p^*)$, which is larger than the counterpart obtained by the cross-entropy loss when $\tau_{\max}> 1$. Our theoretical analysis suggests that the adaptive scaling factor essentially changes the correspondence between the logit of preference distribution and the reward difference for each pair of trajectory segments, which could lead to a more flexible reward model.

\begin{figure}[htb!]
%\vspace{-0.10in}
\centering
\subfigure{
\includegraphics[width=0.3\linewidth]{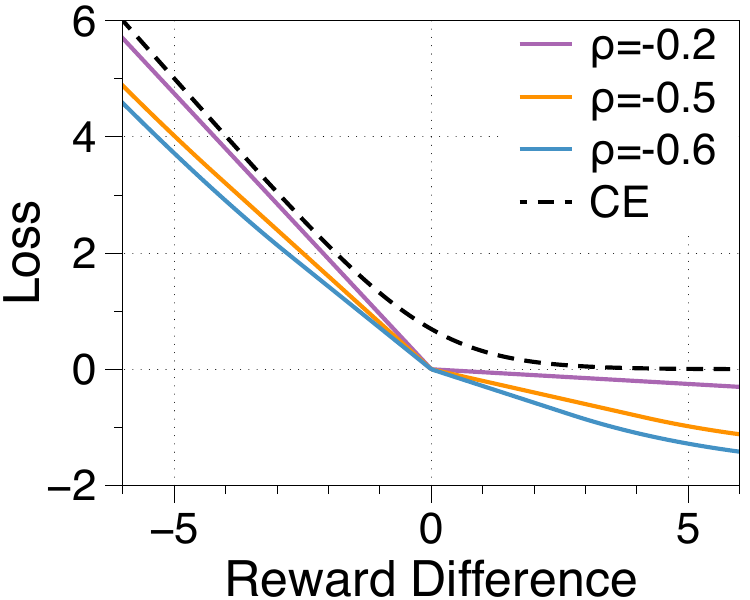}%
}
\hspace{1.5cm}
\subfigure{
\includegraphics[width=0.3\linewidth]{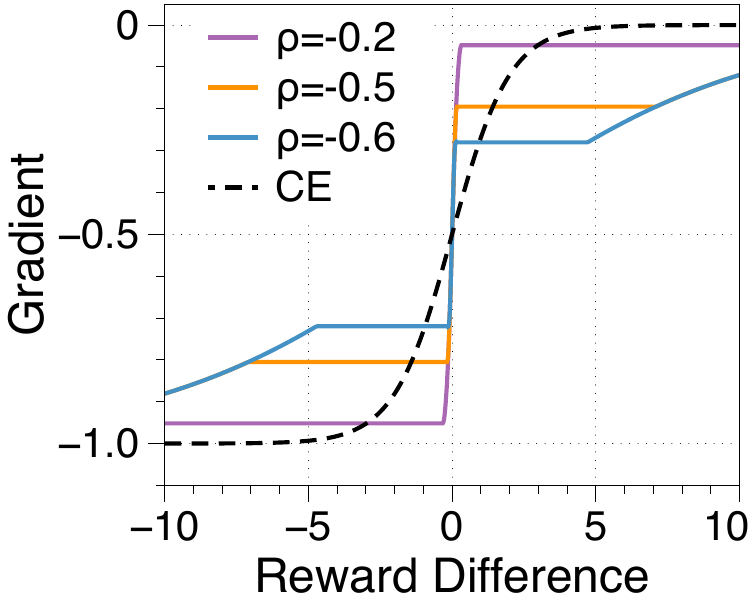}%
}
%%\vspace{-0.35in}
\caption{Visualization of the loss function (left) and its gradient (right) on different reward differences.}
\label{fig:visual_loss}
%\vspace{-0.05in}
\end{figure}

We further visualize our adaptive preference loss in Figure \ref{fig:visual_loss}, setting $\tau_0$ and $\tau_{\rm max}$ to $0.1$ and $5.0$, respectively. As depicted, our adaptive preference loss behaves distinctly compared to the cross-entropy loss. With large learned reward differences, the cross-entropy tends to be very flat, while our loss maintains a non-trivial gradient, allowing us to continually decrease the loss function. In contrast, for small positive learned reward differences, our loss yields a smaller gradient, thereby less encouraging the reward model to further distinguish pairs of ambiguous trajectory segments. This is consistent with our theoretical analysis.

\subsection{Algorithm}

We present an efficient algorithm for solving \eqref{eq:adap-rlhf}. Suppose we parameterize $r$ as a neural network with parameter $\phi$. At the $m$-th iteration, we have the iterate $\phi^{(m)}$, and we sample a pair of trajectory segments $z_{w,i}$ and $z_{l,i}$. We initialize $\tau_i^{(0)}=1$ and then optimize $\tau_i$ by a projected Newton method subject to a simple interval constraint $\Omega$ \cite{bertsekas1982projected}. Specifically, for $k=0,...,K-1$, we take
\begin{align}\label{eq:projected_newton}
\tau_i^{(k+1)} = \underset{\tau_i\in\Omega}{\Pi}(\tau_i^{(k)} + \Delta_i^{(k)}),
\end{align}
where $\Delta_i^{(k)}$ denotes the descent direction
\begin{align}\label{eq:newton-step}
\Delta_i^{(k)}=-\frac{\nabla_{\tau_i}\ell_i(\phi^{(m)},\tau_i^{(k)})}{\nabla^2_{\tau_i}\ell_i(\phi^{(m)},\tau_i^{(k)})}.
\end{align}
Once we get $\tau_i^{(K)}$, we update $\phi$ by a stochastic gradient descent step
\begin{align}\label{phi-SGD}
\phi^{(m+1)} = \phi^{(m)} - \eta_{\phi}\nabla_{\phi}\ell_i(\phi^{(m)},\tau_i^{(K)}),
\end{align}
where $\eta_{\phi}$ is the learning rate.
We summarized our proposed algorithm in Algorithm~\ref{alg:alg1}.

\begin{algorithm}[htp!]
\caption{Algorithm for reward learning with adaptive preference scaling}\label{alg:alg1}
\begin{algorithmic}[1]
\STATE{\bfseries Input}: $\tau_0$, $\tau_{\max}$, $\rho$, $\eta_{\phi}$;
\FOR{$m=0,1,2,\dots, M-1$}
\STATE Sample a pair of trajectory segements from $\cD_{\mathrm{pref}}$;
\STATE Set $\tau_i^{0}=1$;
\FOR{$k=0,1,2,\dots, K-1$}
\STATE Compute $\Delta_i^{(k)}$ using \eqref{eq:newton-step} and update $\tau_i^{(k)}$ using \eqref{eq:projected_newton};
\ENDFOR
\STATE Update $\phi^{(m)}$ using \eqref{phi-SGD} or Adam-style step;
\ENDFOR
\end{algorithmic}
\end{algorithm}

\begin{remark}
Note that since $\ell_i(\phi,\tau_i)$ is strictly convex and univariate with respect to $\tau_i$, in each iteration $m$, $\tau^{(K)}_i$ is guaranteed to be near-optimal (i.e., $\tau^{(K)}_i\approx\tau_i^\star$). Therefore, the convergence of Algorithm \ref{alg:alg1} can be guaranteed by the convergence of stochastic gradient descent on the reward model parameter $\phi$.
\end{remark}

% Since $\ell_i(\phi,\tau_i)$ is strictly convex, differentiable, and univariate with respect to $\tau_i$, in each iteration $m$, we are guaranteed to find $\tilde{\tau}_i$ such that $|\tilde{\tau}_i-\tau^\star_i|\le\delta$ for any $\delta>0$. Under the assumptions that $r_{\phi}(z_{w,i})-r_{\phi}(z_{l,i})$ is $L$-smooth, $G$-Lipschitz continuous, and bounded by $C$, which are standard for convergence analysis in KL-constrained DRO literatures~\cite{qi2022stochastic, qiu2023not}, we have
% \begin{align*}
% \|\nabla_{\phi}\ell_i(\phi,\tilde{\tau}_i)-\nabla_{\phi}\ell_i(\phi,\tau^\star_i)\|&\le\|\nabla_{\phi}(r_{\phi}(\traj_{w,i})-r_{\phi}(\traj_{l,i}))\|\|\sigmoid((r_{\phi}(\traj_{w,i})-r_{\phi}(\traj_{l,i}))/\tilde{\tau}_i)-\sigmoid((r_{\phi}(\traj_{w,i})-r_{\phi}(\traj_{l,i}))/\tau^\star_i)\|\\
% &\le G\max_{\tau\in [\tilde{\tau}_i, \tau^\star_i]}|\sigmoid'((r_{\phi}(\traj_{w,i})-r_{\phi}(\traj_{l,i}))/\tau)\cdot(r_{\phi}(\traj_{w,i})-r_{\phi}(\traj_{l,i}))/\tau^2)|\cdot|\tilde{\tau}_i-\tau^\star_i|\\
% &\le \frac{CG}{4\tau_0^2}\delta,
% \end{align*}
% where the second inequality is due to the mean value theorem. Therefore, the convergence of Algorithm \ref{alg:alg1} can be guaranteed by the convergence of stochastic gradient descent.

%Although solving \eqref{eq:adap-rlhf} rigorously requires more complex design such as a line search in Newton's method, we found that our simple algorithm effectively captures the advantages in our adaptive preference loss in practical settings.

\subsection{Extension to Direct Preference Optimization (DPO)}

Our adaptive preference scaling approach is generic and can be extended to DPO~\citep{rafailov2023direct}, which is another popular method for policy learning from human preferences. DPO directly learns the policy in supervised manner using the preference data of state-action pairs $\cD_{\mathrm{pref}}=(s_i,a_{w,i}, a_{l,i})_{i=1}^N$. This approach forgoes the need to learn the reward function explicitly by the reparameterization of reward function $r$ with respect to its optimal policy $\pi_r$,
\begin{align}\label{eq:reparameterization}
r(s,a)=\beta\log\dfrac{\pi_r(a|s)}{\pi_{\mathrm{ref}}(a|s)}+\beta\log Z(s),
\end{align}
where $Z(s)=\sum_{a}\pi_{\mathrm{ref}}(a|s)\exp\big(r(s,a)/\beta\big)$ and $\pi_{\mathrm{ref}}$ denotes the reference policy. By plugging in \eqref{eq:reparameterization} back into \eqref{eq:rm}, we have the policy optimization problem
% \begin{align}\label{eq:dpo}
% &\underset{\theta}{\min}\;\mathcal{L}(\theta)\nonumber\\
% &=-\mathbb{E}_{(s,a_{w}, a_{l})\sim\gD_{\text{pref}}}\Bigg[\log\sigma\bigg(\beta\log\dfrac{\pi_{\theta}(a_{w}|s)}{\pi_{\text{ref}}(a_{w}|s)}-\beta\log\dfrac{\pi_{\theta}(a_{l}|s)}{\pi_{\text{ref}}(a_{l}|s)}\bigg)\Bigg].
% \end{align}
\begin{align*}
\underset{\pi}{\min}\;\mathcal{L}_{\mathrm{DPO}}(\pi)=-\mathbb{E}_{(s,a_{w}, a_{l})\sim\cD_{\mathrm{pref}}}\log\sigma\big(\beta r_{\pi}(a_w|s)-\beta r_{\pi}(a_l|s)\big),
\end{align*}
where $r_{\pi}(a|s) = \log(\pi(a|s)/\pi_{\mathrm{ref}}(a|s))$ denotes the log-probability ratio. 
%where $r_{\pi}(a|s) = \log\dfrac{\pi(a|s)}{\pi_{\mathrm{ref}}(a|s)}$ denotes the log-probability ratio. %Note that compared to RLHF, DPO relies only on the pre-collected preference data while not using policy-generated, preference-free data.

% \textbf{DPO with adaptive temperature scaling.} 
Similarly. we can integrate adaptive preference scaling into DPO by plugging in \eqref{eq:reparameterization} into \eqref{eq:adap-rlhf}.
By merging $\beta$ with the $\tau_i$ and $\rho$, we can further obtain the adaptive DPO (Ada-DPO) formulation as
\begin{align*}
\underset{\pi, \tau_1,...,\tau_N\in\Omega}{\min}\;\mathcal{L}_{\mathrm{Ada-DPO}}(\pi,\tau_1,...,\tau_N):=\frac{1}{N}\sum_{i=1}^N\bigg[-\tau_i\log\sigma\bigg(\dfrac{r_{\pi}(a_{w,i}|{s_i})- r_{\pi}(a_{l,i}|s_i)}{\tau_i}\bigg)+\rho\tau_i\bigg].
\end{align*}

\begin{remark}
Note that the proposed adaptive preference loss can be further combined with other RLHF approaches, such as PEBBLE~\citep{lee2021pebble} and SURF~\citep{park2022surf}, which still optimize the standard cross-entropy loss (see \citet[Eq. (4)]{lee2021pebble} and \citet[Eq. (3)]{park2022surf}).
\end{remark}
%where $\rho > -\log 2$.
% For DPO formulation, instead of preference over trajectory segments, we consider preference over a pair of actions from a given state. From \citet{rafailov2023direct}, we have the reparameterization of reward function $r$ with respect to its optimal policy $\pi_r$ of \eqref{eq:policy} as
% \begin{align}\label{eq:reparameterization}
% r(s,a)=\beta\log\dfrac{\pi_{r}(a|s)}{\pi_{\text{ref}}(a|s)}+\beta\log Z(s),
% \end{align}
% where $Z(s)=\sum_{a}\pi_{\text{ref}}(a|s)\exp\big(r(s,a)/\beta\big)$. Plugging in \eqref{eq:reparameterization} back into \eqref{eq:temp-BT} and similarly with \eqref{eq:adap-rlhf}, we have our policy objective with adaptive temperature scaling for a given dataset $\gD=(s_i,a_{w,i}, a_{l,i})_{i=1}^N$
% \begin{align}\label{eq:dpo}
% \resizebox{0.99\columnwidth}{!}{$\mathcal{L}(\theta,\tau):=-\dfrac{1}{N}\sum_{i=1}^N\tau_{i}\log\Bigg(\sigma\bigg(\dfrac{\beta}{\tau_{i}}\log\dfrac{\pi_{\theta}(a_{w,i}|s_i)}{\pi_{\text{ref}}(a_{w,i}|s_i)}-\dfrac{\beta}{\tau_{i}}\log\dfrac{\pi_{\theta}(a_{l,i}|s_i)}{\pi_{\text{ref}}(a_{l,i}|s_i)}\bigg)\Bigg)+\rho\bar{\tau}$},
% \end{align}
% where $\bar{\tau}=(1/N)\sum_{i=1}^N\tau_{i}$, and $\rho>-\log2$, $\beta>0$.

\subsection{Extension to Quadratic Regularization}\label{sec:quad}

We now introduce a variant of our adaptive preference loss that uses quadratic regularization for $\tau$. This modification removes the need for the hyperparameter $\tau_{\max}$ in $\Omega$, easing the tuning effort. We define the following instance-level adaptive preference loss with quadratic regularization:
\begin{align}\label{eq:ada-pref2}
\underset{\tau\ge\tau_0}{\min}\;\ell_{\mathrm{quad}}(r,\tau):=-\tau\log p_{r,\tau}(\traj_w\succ \traj_l)+\rho_0\tau^2-\log2\tau.
\end{align}
Compared to \eqref{eq:drpf}, which includes a linear regularization term of $(\rho_0-\log2)\tau$, \eqref{eq:ada-pref2} modifies the regularization term with coefficient $\rho_0$ to be quadratic while keeping the term $\log 2 \tau$ linear. Additionally, in \eqref{eq:ada-pref2}, the constraint on $\tau$ only specifies a lower bound $\tau_0$ and no longer includes an upper bound $\tau_{\max}$. The following proposition provides theoretical insights for this modification.

\begin{proposition}\label{prop:2}
    Assume we have a pair of trajectory segments $z_1, z_2$, and the preference distribution $p(z_1\succ z_2)=p^*\in(0,1)$. Consider the problem of minimizing the expectation of our adaptive loss function with quadratic regularization over the preference distribution:
\begin{align*}
\min_{r,\tau\ge\tau_0}-\tau p^*\log \big(\sigma\big((r(z_1)-r(z_2))/\tau\big)\big)-\tau (1-p^*)\log \big(\sigma\big((r(z_2)-r(z_1))/\tau\big)\big)+\rho_0\tau^2-\log2\tau.
\end{align*}
Then the minimizer $\tau^*$ and $r^*$ of the expected loss satisfy
\begin{align*}
&\tau^\star=\max\{\tau_0,(p^*\log (p^*)+(1-p^*)\log (1-p^*)+\log2)/(2\rho_0)\},\\
&r^*(z_1)-r^*(z_2)=\tau^* \sigma^{-1}(p^*).
\end{align*}
Here, $\sigma^{-1}$ is the inverse of sigmoid function.
%\vspace{-0.1in}
\end{proposition}

Note that unlike the adaptive preference loss with linear regularization described in Proposition \ref{prop:1}, the optimal value $\tau^\star$ for quadratic regularization does not involve the upper bound $\tau_{\max}$.

\section{Experiments}

In this section, we examine the effectiveness of our adaptive preference loss based on robotic control and natural language generation tasks.
% We focus on the adaptive loss with linear regularization and present the results and analysis in the main text. Due to space constraints, we provide the results for the extension with quadratic regularization in the Appendix~\ref{sec:appen_exp2_quad}.

\subsection{Robotic Control}

\begin{figure}[!b]
\centering
%\subfigure{
\begin{minipage}[t]{0.26\linewidth}
\includegraphics[width=0.97\linewidth]
{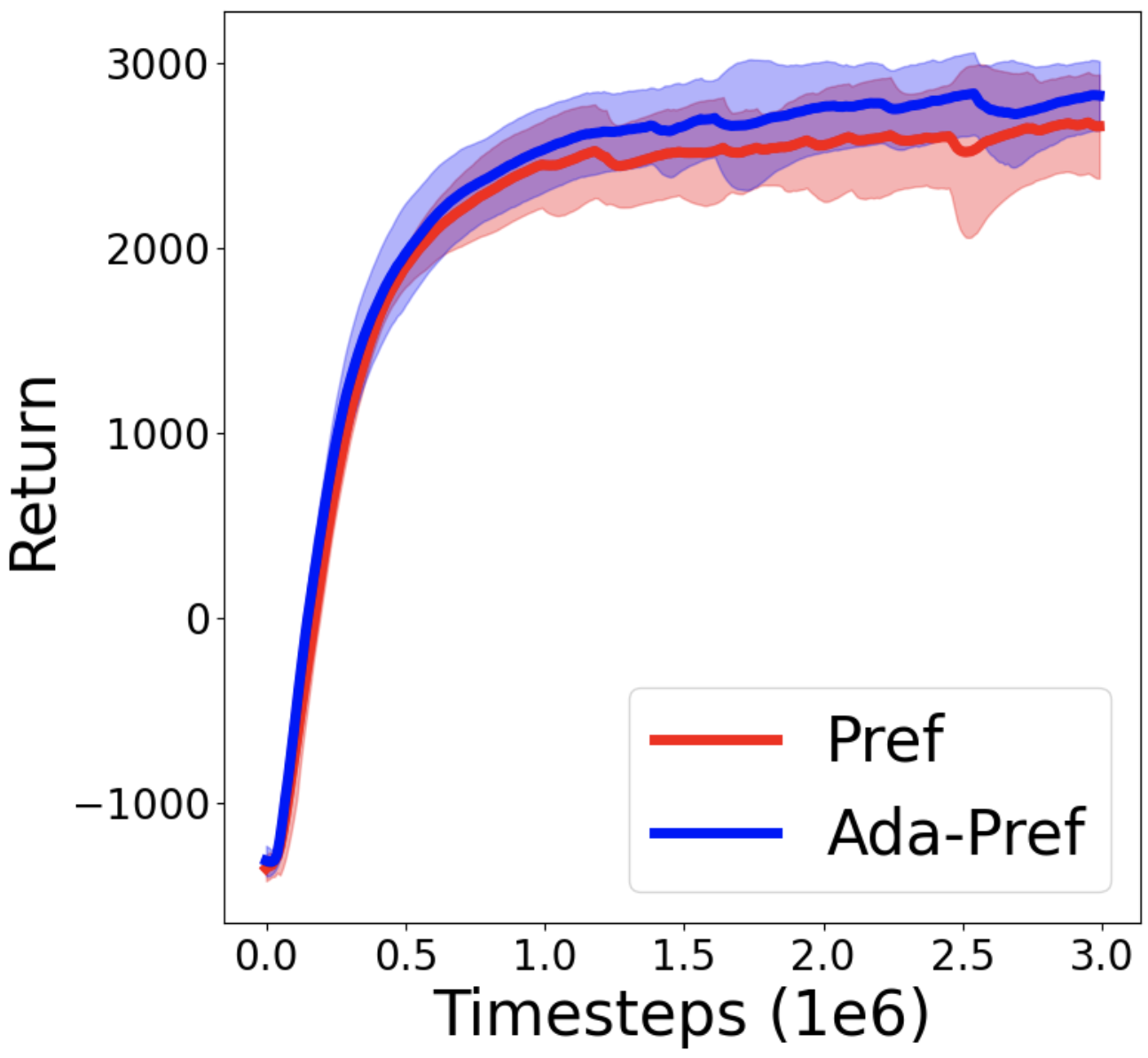}
\end{minipage}
%}
%\subfigure{
\begin{minipage}[t]{0.26\linewidth}
\includegraphics[width=0.97\linewidth]
{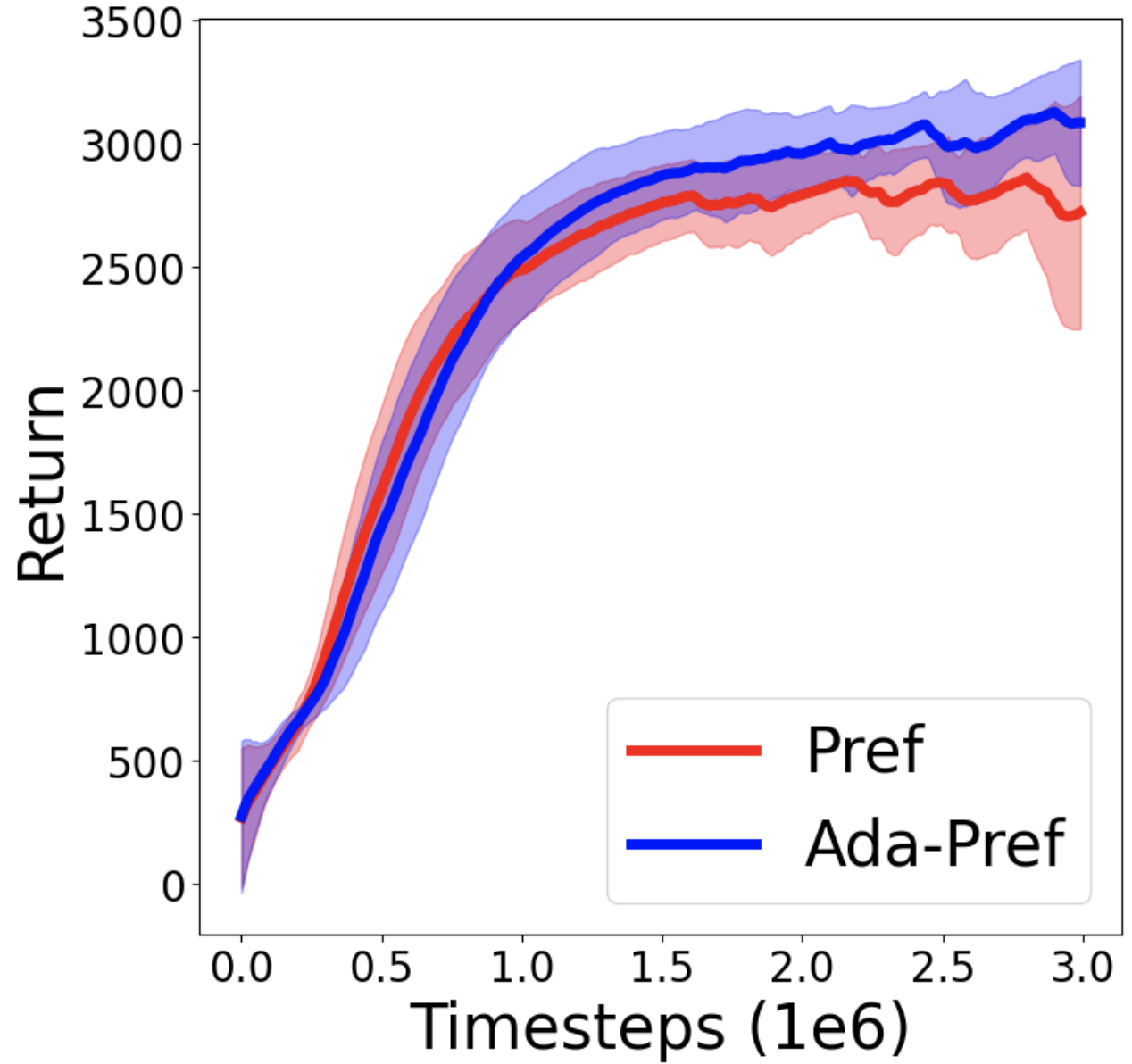}
\end{minipage}
%}
%\subfigure{
\begin{minipage}[t]{0.26\linewidth}
\includegraphics[width=0.97\linewidth]
{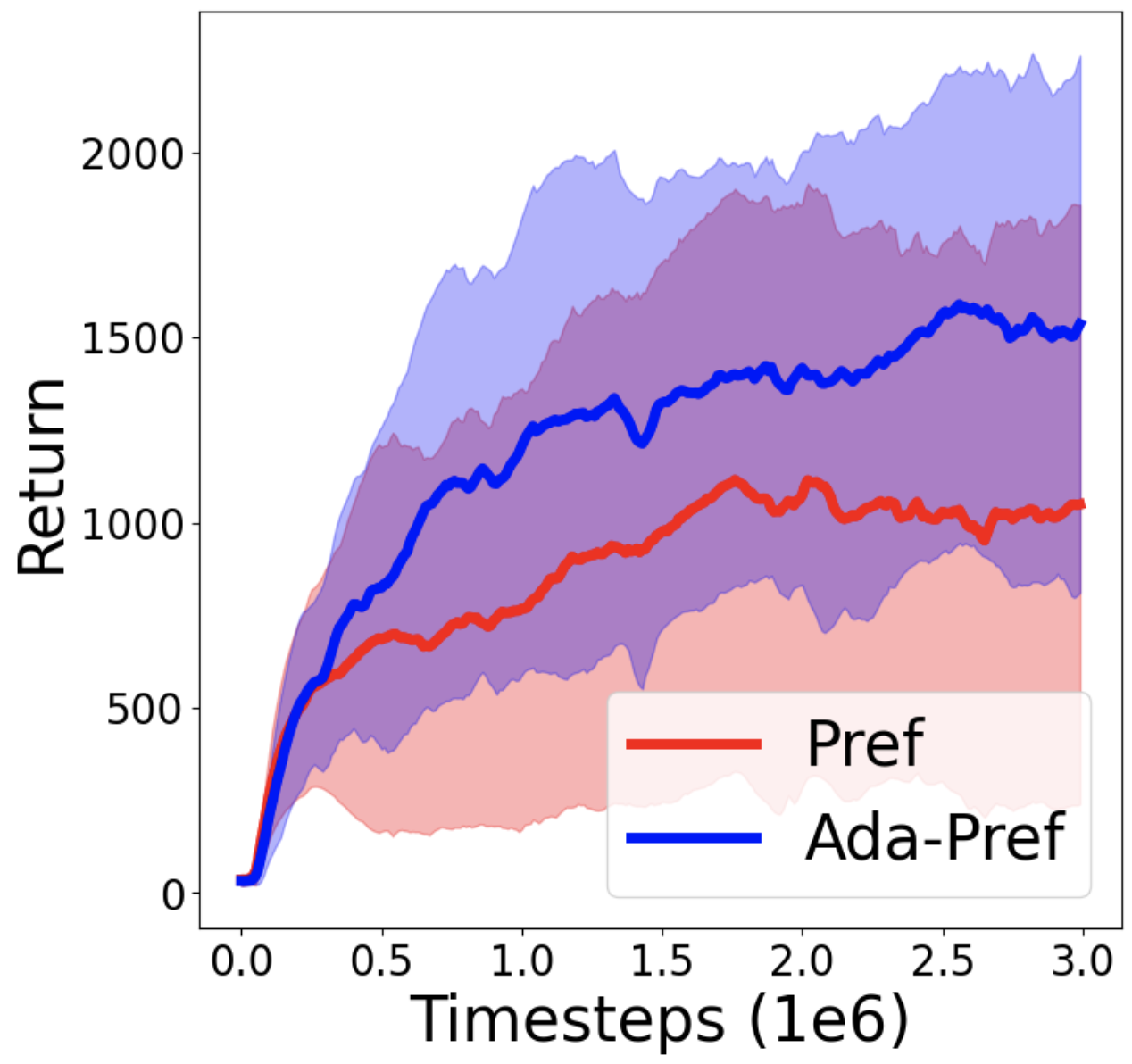}
\end{minipage}
%}
\\
\subfigure[HalfCheetah]{
\includegraphics[width=0.25\linewidth]
{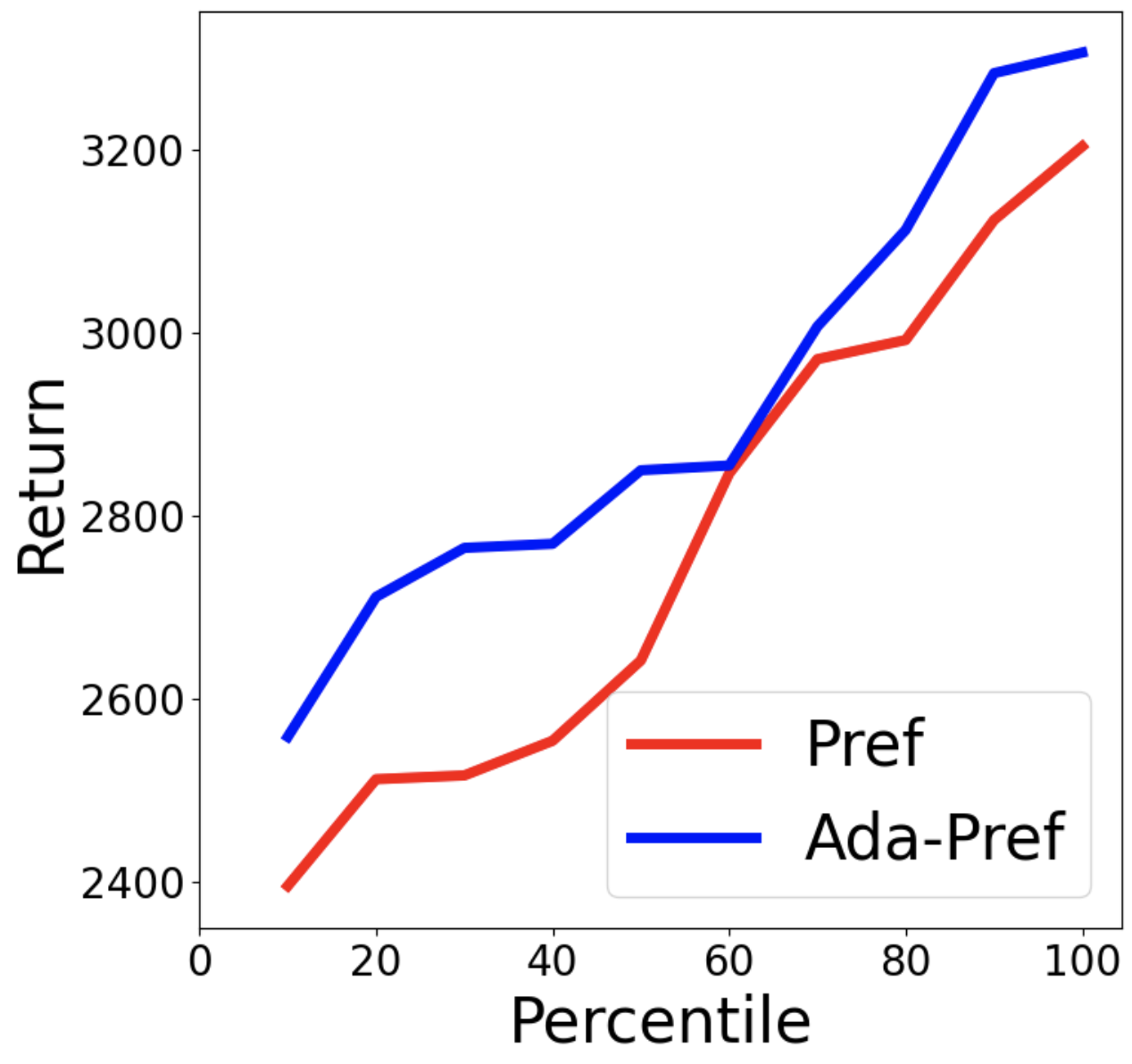}
}
\subfigure[Ant]{
\includegraphics[width=0.25\linewidth]
{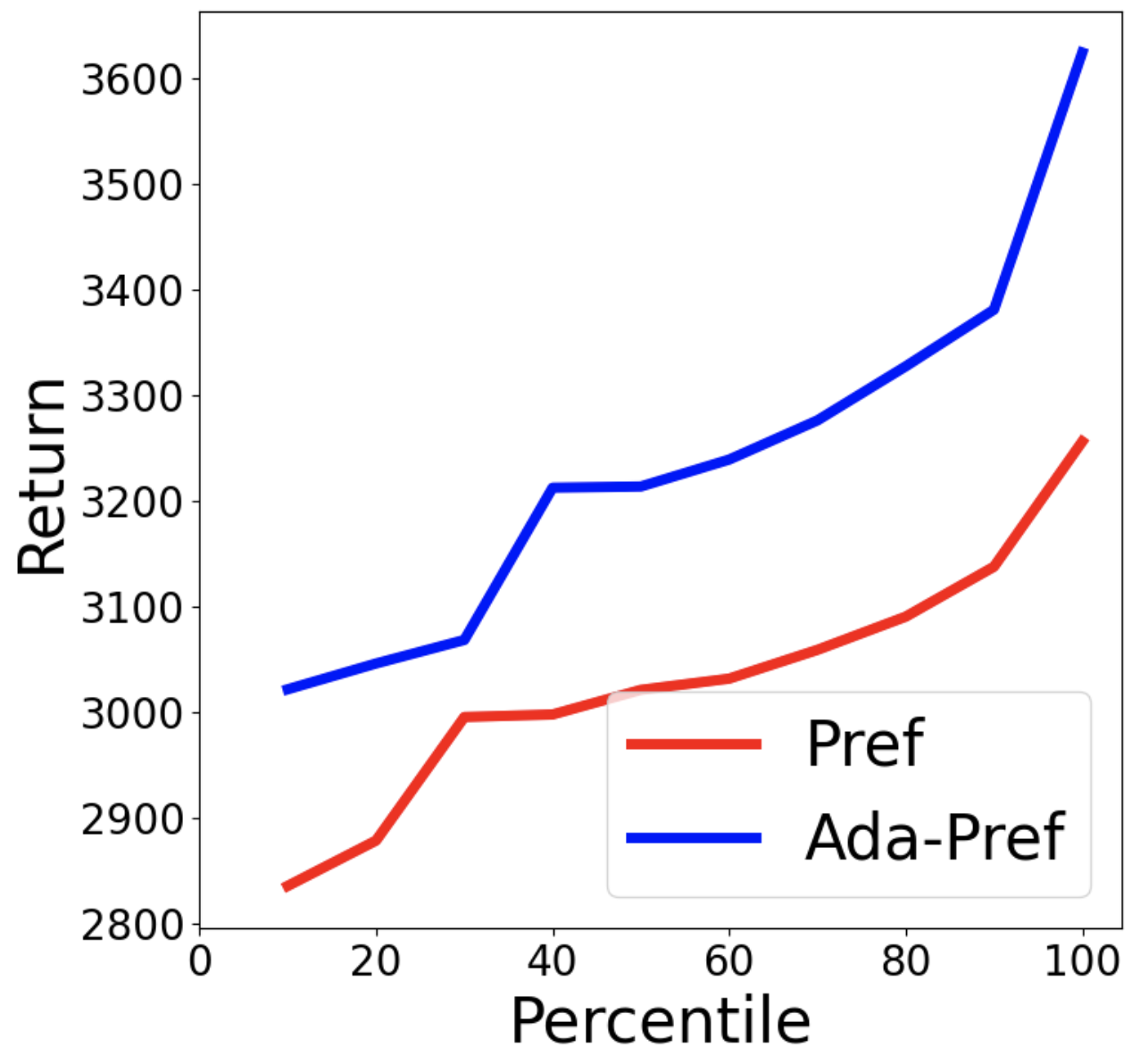}
}
\subfigure[Hopper]{
\includegraphics[width=0.25\linewidth]
{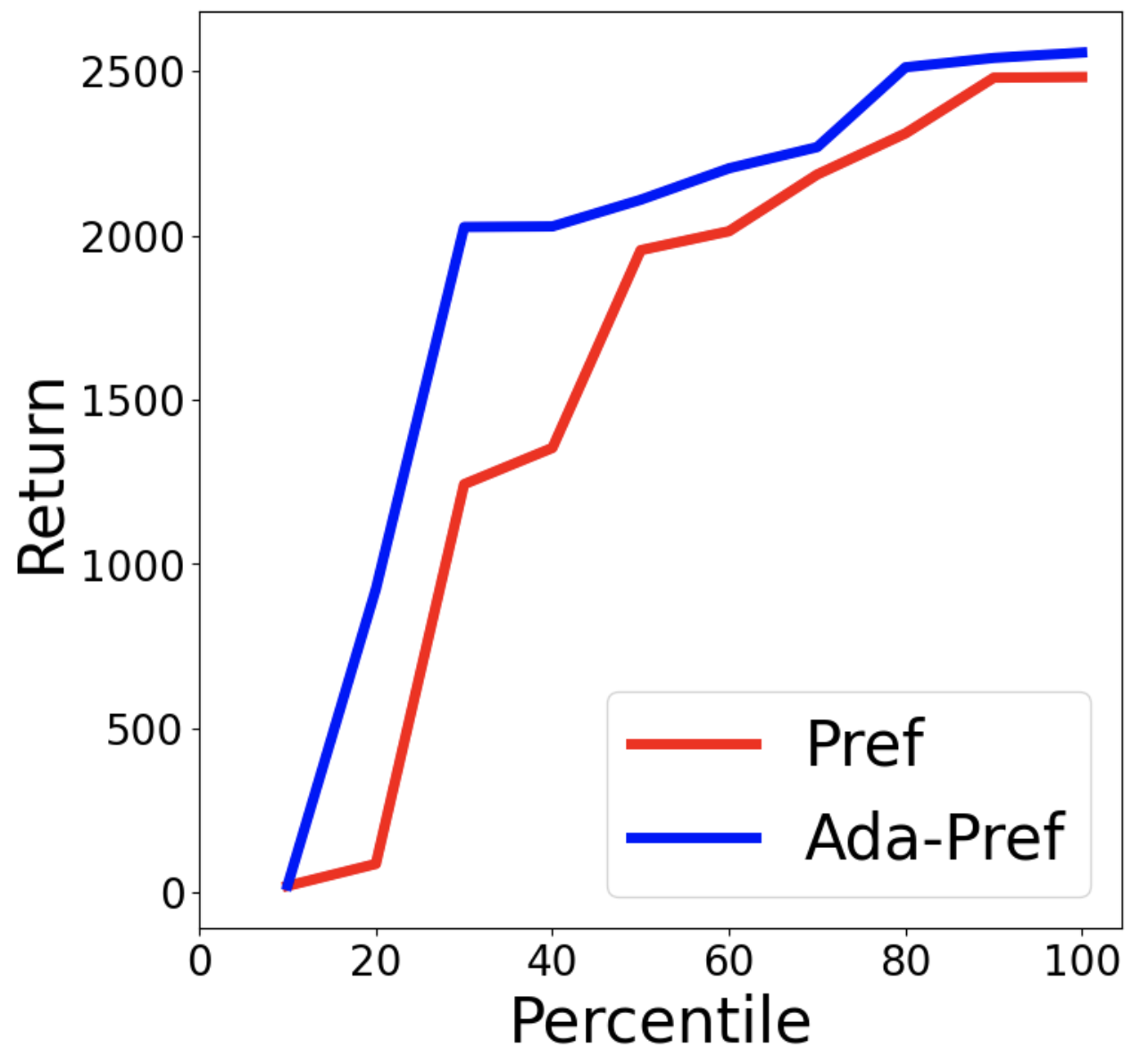}
}
\caption{Learning curve plots (top) and percentile plots (bottom) for Pref and Ada-Pref. For the learning curve plots, returns at each timestep are averaged across 10 different seeds, then smoothed over timesteps using an exponential moving average (EMA) with a smoothing factor of $\alpha=0.1$. For the percentile plots, returns from 10 different seeds are sorted in ascending order.
}
\label{fig:pybullet_exp}
\end{figure}

% \begin{figure}[htb!]
% \centering
% \subfigure{
% \includegraphics[width=0.31\linewidth]
% {Figures/half_curve_q.png}
% }
% \subfigure{
% \includegraphics[width=0.31\linewidth]
% {Figures/ant_curve_q.png}
% }
% \subfigure{
% \includegraphics[width=0.31\linewidth]
% {Figures/hopper_curve_q.png}
% }
% \\
% \subfigure[HalfCheetah]{
% \includegraphics[width=0.31\linewidth]
% {Figures/half_percentile_q.png}
% }
% \subfigure[Ant]{
% \includegraphics[width=0.31\linewidth]
% {Figures/ant_percentile_q.png}
% }
% \subfigure[Hopper]{
% \includegraphics[width=0.31\linewidth]
% {Figures/hopper_percentile_q.png}
% }
% \caption{Learning curve plots (top) and percentile plots (bottom) for Ada-Pref and Pref. For the learning curve plots, returns at each timestep are averaged across 10 different seeds, then smoothed over timesteps using an exponential moving average (EMA) with a smoothing factor of $\alpha=0.1$. For the percentile plots, returns from 10 different seeds are sorted in ascending order.
% }
% \label{fig:pybullet_exp}
% \end{figure}

\textbf{Experiment Setup.} We apply our proposed reward learning method on 3 robotic control tasks from the PyBullet~\citep{pybullet} environments: \textit{HalfCheetah}, \textit{Ant}, and \textit{Hopper}. These environments are similar to those available in OpenAI Gym \citep{brockman2016openai} but they are known to be much harder to solve \citep{raffin2022smooth}. Similarly to \citet{gao2023scaling}, our setting is synthetic, where we use the ground truth rewards to provide preference labels on each pair of samples due to high expense of collecting human preferences. For the reward function, we use two-hidden-layer MLPs, each containing 64 hidden units. This configuration is aligned with the designs of both the policy and value networks. Following \citet{christiano2017deep}, we repeat the following three steps for each stage: (i) We sample a set of trajectories by the policy $\pi$, and update the policy with proximal policy optimization (PPO, \citet{schulman2017proximal}) alongside a reward function $\hat{r}$.
(ii) We split the segments (the sequence of state-action pairs) into a training set and a testing set. Then, we randomly sample pairs of segments from the training set, and generate $\cD_{\mathrm{pref}}$ with preference labels. We do the same to the testing set, and generate $\cD_{\mathrm{pref}}'$.
(iii) We train the reward function $\hat{r}$ on $\cD_{\mathrm{pref}}$, and use $\cD_{\mathrm{pref}}'$ for evaluating the preference prediction of $\hat{r}$.

% Following \citet{christiano2017deep}, we repeat the following three steps for each stage: 

% Step 1. We sample a set of trajectories by the policy $\pi$, and update the policy with proximal policy optimization (PPO, \citet{schulman2017proximal}) alongside a reward function $\hat{r}$.

% %%%We compute the cumulative ground truth rewards of $\pi$ based on the sampled trajectory.%%%

% Step 2. We split the segments (the sequence of state-action pairs) into a training set and a testing set. Then, we randomly sample pairs of segments from the training set, and generate $\cD_{\mathrm{pref}}$ with preference labels. We do the same to the testing set, and generate $\cD_{\mathrm{pref}}'$.

% Step 3. We train the reward function $\hat{r}$ on $\cD_{\mathrm{pref}}$, and use $\cD_{\mathrm{pref}}'$ for evaluating the preference prediction of $\hat{r}$.

% To evaluate the learned policy, we compute the cumulative ground truth rewards of $\pi$ based on 5 evaluation episodes. 
For notational simplicity, we name our adaptive preference scaling method for reward learning as ``Ada-Pref''. We compare Ada-Pref with the baseline method ``Pref'', which uses the standard cross-entropy loss for reward learning. For every 10000 timesteps the policy $\pi$ runs, we evaluate the learned policy based on 20 test episodes. We also compute the average preference prediction accuracy of the learned reward function across stages. We set the budget to 3 million timesteps and perform training over 10 different seeds. For hyperparameter tuning in both reward learning and policy optimization, we apply two different criteria: 1) We identify the best policy based on its performance (the one with the highest return) and subsequently select the corresponding reward function. 2) We choose the best reward function based on its performance (the one with the highest average preference prediction accuracy) and then select the corresponding policy. Details of the implementations and hyperparameter tuning procedures are in Appendix \ref{sec:appen_exp:robot}.

\textbf{Results.} We summarize the results on three PyBullet tasks as follows:

Table \ref{tab:pybullet1} and Figure \ref{fig:pybullet_exp} illustrate the results for Pref and Ada-Pref on the PyBullet tasks, based on the first hyperparameter tuning criterion. In Table \ref{tab:pybullet1}, we report the highest return of the best policy and the average preference accuracy of the corresponding reward function. We can see that Ada-Pref consistently outperforms Pref in terms of return on all three tasks and achieves comparable preference accuracy. The upper panel of Figure \ref{fig:pybullet_exp} shows the learning curve plots. We can see that Ada-Pref surpasses Pref at nearly every timestep and reaches a higher plateau across all tasks. The lower panel of Figure~\ref{fig:pybullet_exp} presents percentile plots from different seeds to demonstrate individual run behaviors. As shown, we confirm that Ada-Pref consistently outperforms Pref at every percentile across all tasks.

Table \ref{tab:pybullet2} presents the results for Pref and Ada-Pref based on the second hyperparameter tuning criterion. In Table \ref{tab:pybullet2}, we report the average preference accuracy of the best learned reward function and the highest return of the corresponding policy. From Table \ref{tab:pybullet2}, we can see that both methods show a decrease in performance compared to Table \ref{tab:pybullet1}, while Ada-Pref still outperforms Pref in terms of both preference accuracy and return on all three tasks. Furthermore, Ada-Pref demonstrates greater resistance to performance degradation than Pref, indicating its superior ability to align the learned reward function with policy optimization. This alignment allows for effective policy selection based on preference accuracy without the need to evaluate the policy using ground truth rewards.
% It also includes the percentage decrease in the highest return compared to those reported in Table \ref{tab:pybullet1}. 

% \begin{table}[htb!]
% \caption{Table for the average preference accuracy of the best reward function and the highest return of the corresponding policy.}
% \label{tab:pybullet2}
% \centering
% \begin{tabular}{clcc}
% \toprule
% \multirow{2}{*}{\bf Task}  & \multirow{2}{*}{\bf Method}  & \multirow{2}{*}{\bf Return} & {\bf Preference}   \\ 
% ~ & ~ & ~ & {\bf Accuracy (\%)}  \\\midrule
% \multicolumn{1}{c|}{\multirow{2}{*}{HalfCheetah}}   & \multicolumn{1}{l|}{\baseline}         & 2620.83         & 89.41                \\
% \multicolumn{1}{c|}{}                               & \multicolumn{1}{l|}{\ourmethod}    & \textbf{2865.07}  & 90.75       \\\midrule
% \multicolumn{1}{c|}{\multirow{2}{*}{Ant}}           & \multicolumn{1}{l|}{\baseline}         & 2750.99      & 87.93                \\
% \multicolumn{1}{c|}{}                               & \multicolumn{1}{l|}{\ourmethod}    & \textbf{3008.69}  & 89.23       \\\midrule
% \multicolumn{1}{c|}{\multirow{2}{*}{Hopper}}        & \multicolumn{1}{l|}{\baseline}         & 744.66          & 93.18                \\
% \multicolumn{1}{c|}{}                               & \multicolumn{1}{l|}{\ourmethod}    & \textbf{1134.73}  & 93.26       \\
% \bottomrule
% \end{tabular}
% \end{table}

\begin{table}[htb!]
\centering
\resizebox{.99\linewidth}{!}{%
\begin{minipage}{.48\linewidth}
\caption{Table for the highest return of the best policy and the average preference prediction accuracy of the corresponding reward function.}
\label{tab:pybullet1}
\begin{tabular}{clcc}
\toprule
\multirow{2}{*}{\bf Task}  & \multirow{2}{*}{\bf Method}  & \multirow{2}{*}{\bf Return}  & {\bf Preference}   \\ 
~ & ~ & ~ & {\bf Accuracy (\%)}  \\\midrule
\multicolumn{1}{c|}{\multirow{2}{*}{HalfCheetah}}   & \multicolumn{1}{l|}{\baseline}         & 2724.42           & 89.09                \\
\multicolumn{1}{c|}{}                               & \multicolumn{1}{l|}{\ourmethod}    & \textbf{2875.45}  & 89.46       \\\midrule
\multicolumn{1}{c|}{\multirow{2}{*}{Ant}}           & \multicolumn{1}{l|}{\baseline}         & 2917.81           & 85.57                \\
\multicolumn{1}{c|}{}                               & \multicolumn{1}{l|}{\ourmethod}    & \textbf{3177.11}  & 85.48       \\\midrule
\multicolumn{1}{c|}{\multirow{2}{*}{Hopper}}        & \multicolumn{1}{l|}{\baseline}         & 1324.91           & 92.08                \\
\multicolumn{1}{c|}{}                               & \multicolumn{1}{l|}{\ourmethod}    & \textbf{1692.1}  & 91.36       \\
\bottomrule
\end{tabular}
\end{minipage}
\hspace{0.5cm}
\begin{minipage}{.48\linewidth}
\centering
\caption{Table for the average preference prediction accuracy of the best reward function and the highest return of the corresponding policy.}
\label{tab:pybullet2}
\begin{tabular}{clcc}
\toprule
\multirow{2}{*}{\bf Task}  & \multirow{2}{*}{\bf Method}  & \multirow{2}{*}{\bf Return} & {\bf Preference}   \\ 
~ & ~ & ~ & {\bf Accuracy (\%)}  \\\midrule
\multicolumn{1}{c|}{\multirow{2}{*}{HalfCheetah}}   & \multicolumn{1}{l|}{\baseline}         & 2620.83         & 89.41                \\
\multicolumn{1}{c|}{}                               & \multicolumn{1}{l|}{\ourmethod}    & \textbf{2865.07}  & 90.75       \\\midrule
\multicolumn{1}{c|}{\multirow{2}{*}{Ant}}           & \multicolumn{1}{l|}{\baseline}         & 2750.99      & 87.93                \\
\multicolumn{1}{c|}{}                               & \multicolumn{1}{l|}{\ourmethod}    & \textbf{3008.69}  & 89.23       \\\midrule
\multicolumn{1}{c|}{\multirow{2}{*}{Hopper}}        & \multicolumn{1}{l|}{\baseline}         & 744.66          & 93.18                \\
\multicolumn{1}{c|}{}                               & \multicolumn{1}{l|}{\ourmethod}    & \textbf{1134.73}  & 93.26       \\
\bottomrule
\end{tabular}
\end{minipage}%
}
\end{table}

\subsection{Natural Language Generation}

\textbf{Experiment Setup.} We apply DPO with our proposed adaptive preference loss (Ada-DPO) to two open-ended text generation tasks: \textit{summarization} and \textit{single-turn dialogue}. We adopt the Llama-2 7B model \citep{touvron2023llama} as the backbone and conduct instruction tuning on each task to obtain the initial reference models.
For summarization, the policy generates summaries given posts collected from Reddit.
We use the filtered TL;DR summarization dataset \citep{tldr_2017} for instruction tuning, which contains more than 117K Reddit posts, each with a human-written summary.
We apply the human preferences collected by \citet{learn_summ_2020} for preference optimization, where each transcript contains a pair of responses along with a preference label.
%A confidence score is also annotated by the annotators to indicate their confidence when assigning the preference label.
% We split a subset for validation
For single-turn dialogue, the policy responds to various human queries ranging from simple questions to complex demands.
We utilize the Anthropic Helpful and Harmless dialogue preferences dataset \citep{HH_rlhf} for both instruction tuning and preference optimization. 
This dataset contains 170k human-AI dialogues, with each dialogue containing two AI responses and a human preference label.
We use the preferred responses for instruction tuning and the full set of preferences for optimization.
For instruction tuning stage, we fine-tune the entire Llama-2 model. For the alignment stage using Ada-DPO and different baselines, we apply LoRA fine-tuning for computational efficiency concerns, as we need to simultaneously tune multiple hyperparameters. The rank of the LoRA adaptor is $64$.
We consider three baseline methods: DPO \citep{rafailov2023direct}, $\Psi$ Preference Optimization with Identity Mapping (IPO, \citet{ipo_2022}) and Sequence Likelihood Calibration with Human Feedback (SLiC-HF, \citet{hinge_2023}).

As human evaluation is prohibitively expensive, we use Claude 2 \citep{claude}, a proprietary large language model, to automatically evaluate responses based on summary quality and helpfulness/harmlessness for the summarization and dialogue tasks, respectively. Prior work has shown that Claude 2 and GPT-4 can effectively measure a quantitative improvement over the instruction-tuned model \citep{dubois2023alpacafarm}.
We split a small subset from each instruction tuning dataset for testing and calculate the win rate against the instruction-tuned reference model as the evaluation metric. 
The percentage of instances where the response generated by policy A is preferred over policy B is referred to as the win rate of A against B.
% A 50\% win rate indicates that A and B are equally preferred \alex{This is not true, since there can be ties. We evaluate the responses in both orders}.
We also split a subset from each preference optimization dataset to validate the preference prediction accuracy. Details of the implementations and hyperparameter selections are in Appendix \ref{sec:appen_exp:llm}.

\textbf{Results.} We summarize the results on the two natural language generation tasks as follows:

In Figure \ref{fig:dpo-best}, we select the model with the highest win rate and present the win rate and its preference accuracy for all baselines.
We observe that Ada-DPO outperforms the other baselines on both tasks in terms of win rate and achieves comparable preference accuracy.
In Figure \ref{fig:dpo-acc-best}, we display the performance of the model selected with the highest accuracy (not win rate). As shown, Ada-DPO achieves a significant improvement beyond the DPO baseline in terms of win rate and obtains a comparable preference accuracy. This again indicates that Ada-DPO yields better alignment between the learned reward function and policy optimization, allowing good policy selection based on preference accuracy without a proprietary LLM judge.

\begin{table}[htb!]
\centering
\vspace{-0.1in}
\resizebox{.95\linewidth}{!}{%
\begin{minipage}{.5\linewidth}
\vspace{-0.1in}
\subfigure{
\includegraphics[width=0.48\linewidth]{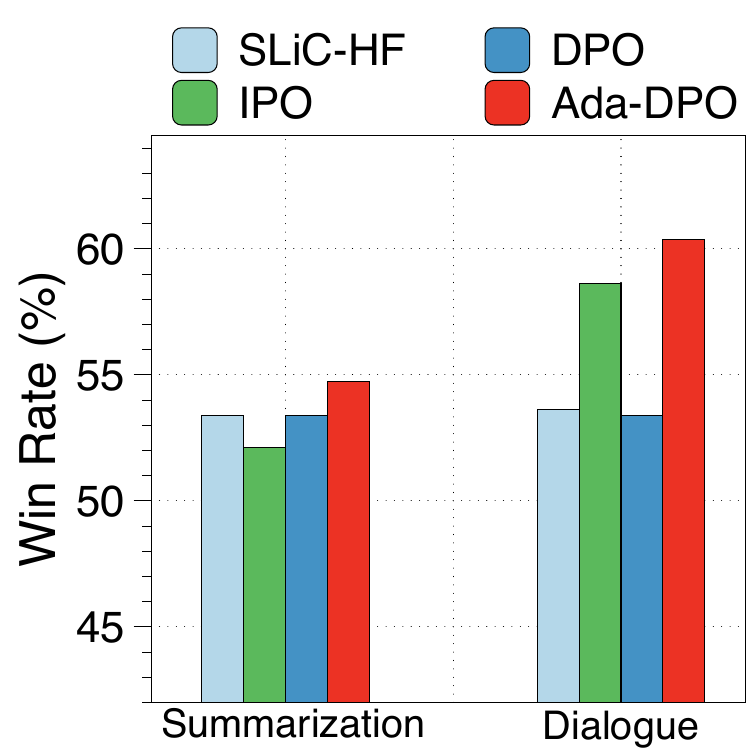}% 
}
\subfigure{
\includegraphics[width=0.48\linewidth]{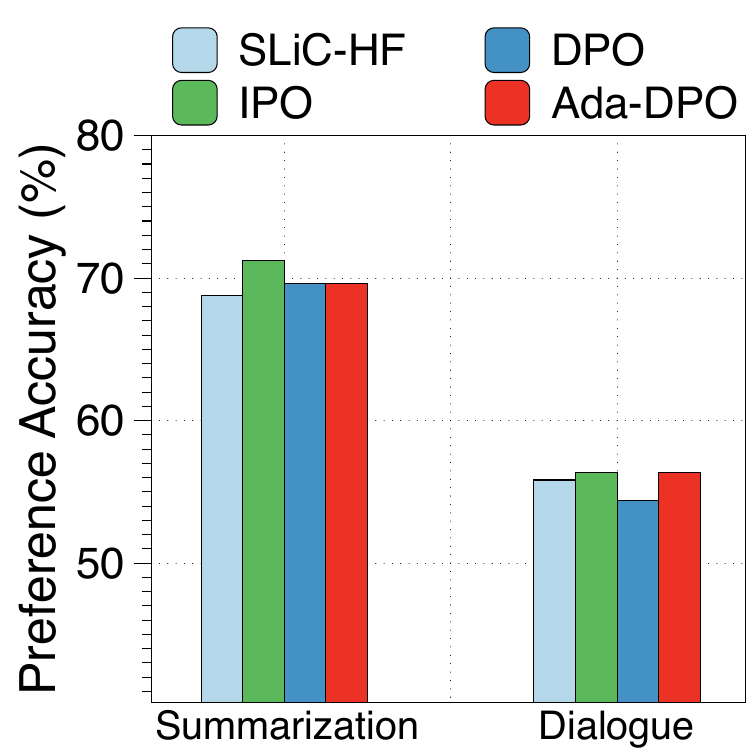}%
}
\captionof{figure}{The best win rate and the preference prediction accuracy of the corresponding model.}
\label{fig:dpo-best}
\end{minipage}
\hspace{0.5cm}
\begin{minipage}{.5\linewidth}
\vspace{-0.1in}
\subfigure{
\includegraphics[width=0.48\linewidth]{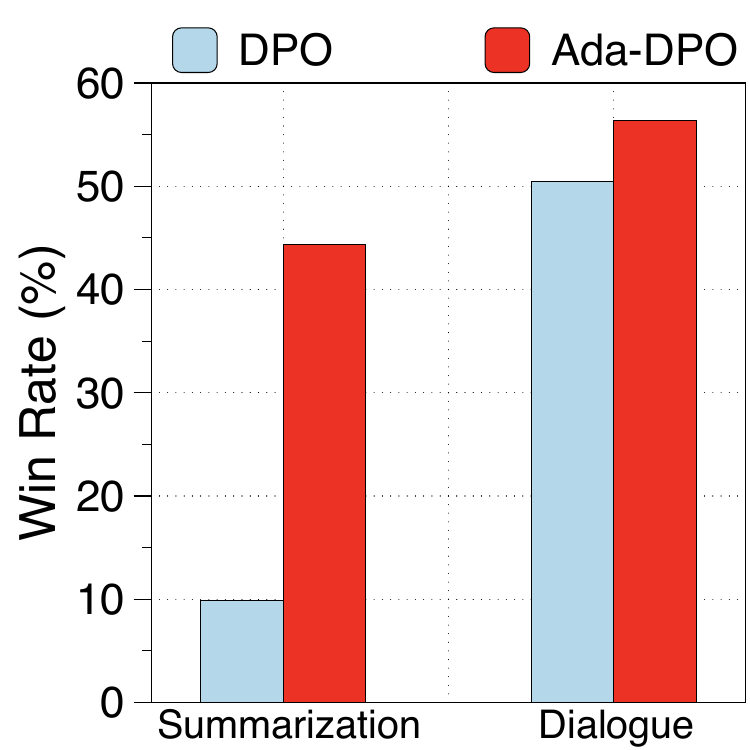}% 
}
\subfigure{
\includegraphics[width=0.48\linewidth]{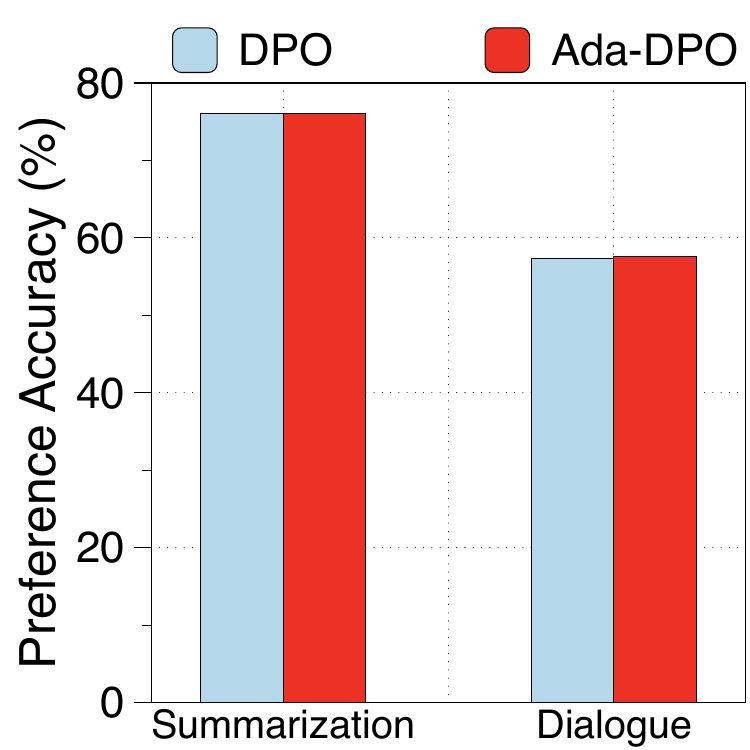}%
}
\captionof{figure}{The best preference prediction accuracy and the win rate of the corresponding model.}
\label{fig:dpo-acc-best}
\end{minipage}%
}
\end{table}

\subsection{Detailed Analysis}

\begin{figure}[!htb]
\centering
\subfigure[Histogram of $\tau$]{\label{fig:ant_tau_dist}
\centering\includegraphics[width=0.28\linewidth]
{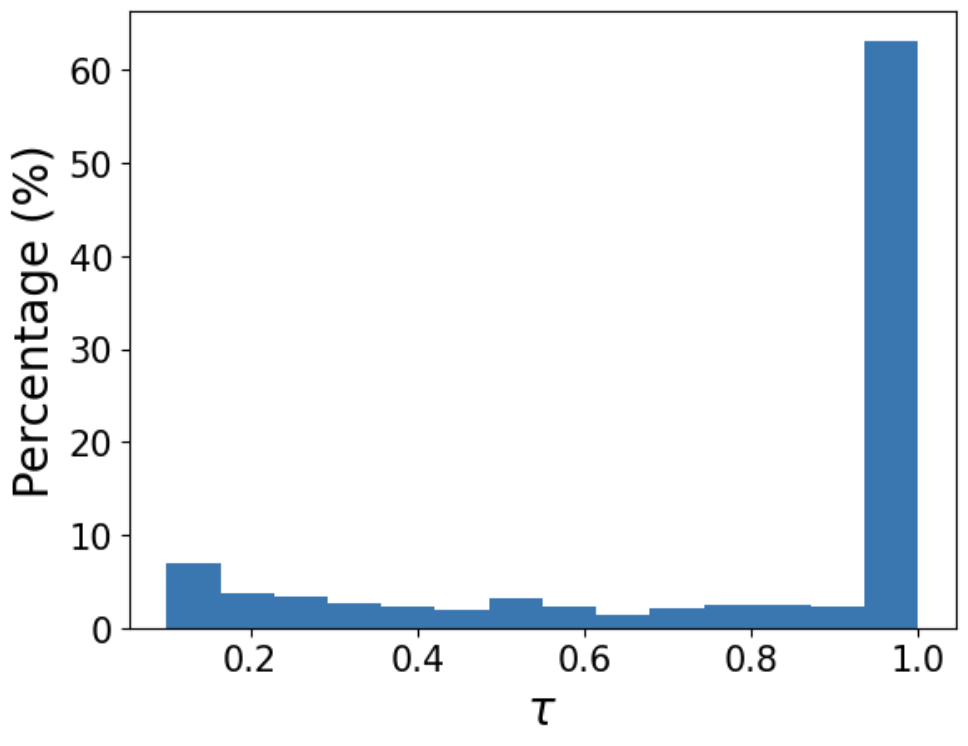}
}
\subfigure[Preference strength and $\tau$]{\label{fig:ant_relation} %
\centering\includegraphics[width=0.28\linewidth]
{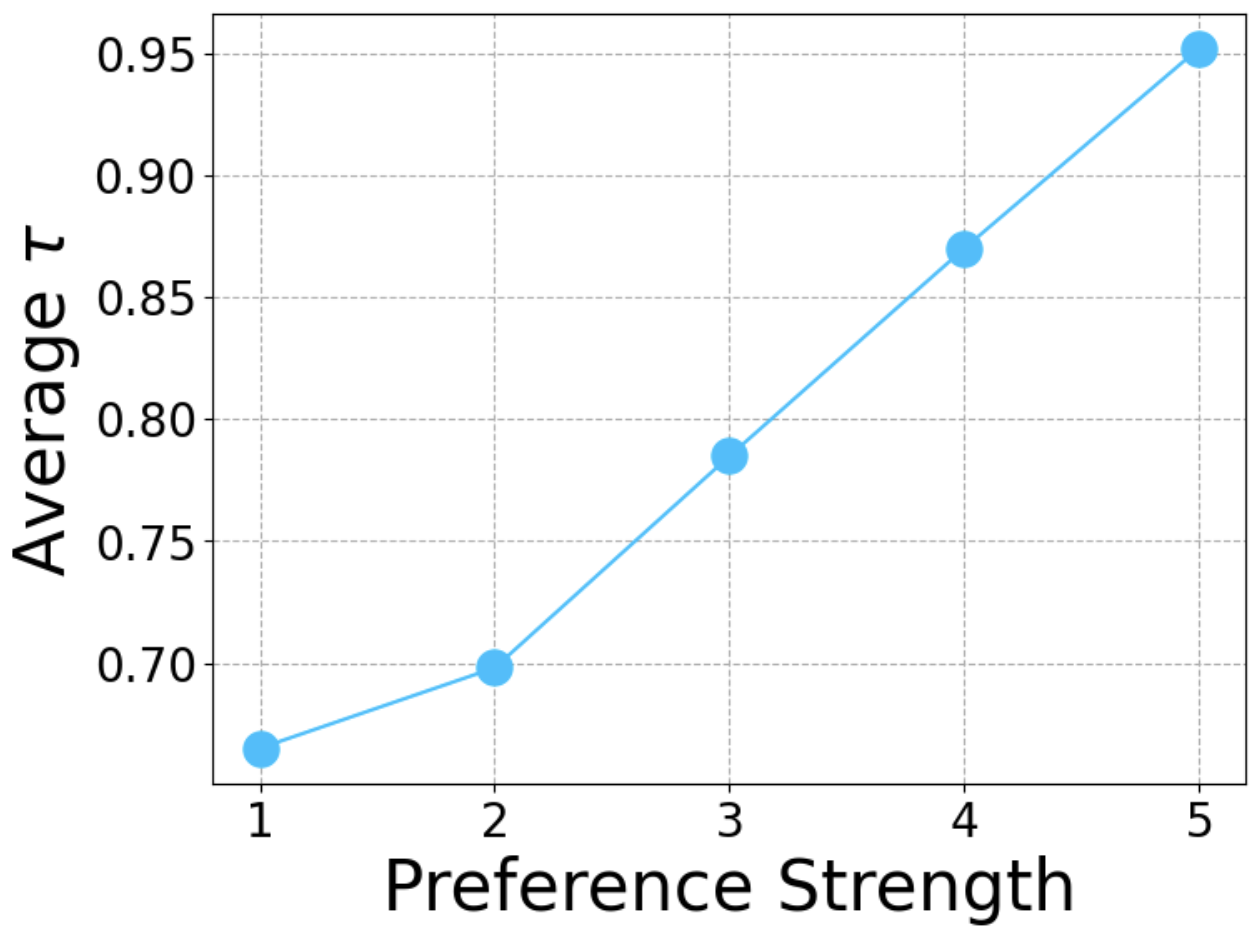}
}
\subfigure[Preference strength and learned reward difference]{\label{fig:flex}
\centering\includegraphics[width=0.28\linewidth]
{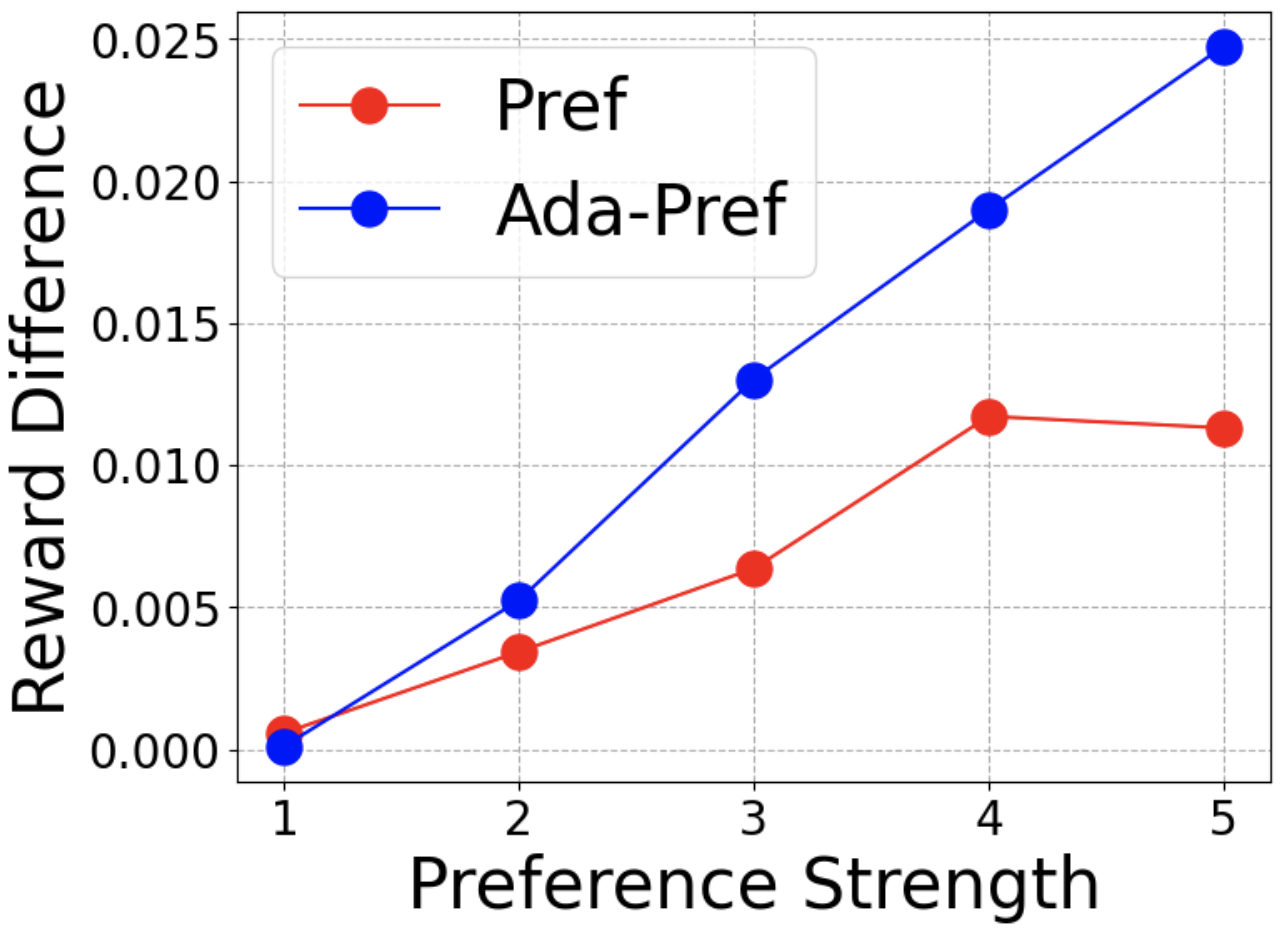}
}
\caption{Histogram of the learned scaling factors, relationship between preference strength and the learned scaling factors, and relationship between preference strength and the learned reward difference. All plots are from the Ant task.}
\end{figure}

We present detailed analyses of Ada-Pref and Ada-DPO for both the Ant and summarization tasks. Figure~\ref{fig:ant_tau_dist} presents a histogram of the learned scaling factors $\tau$ for the Ant task. We can see that around 60\% of these scaling factors reach the upper bound, while about 10\% converge to the lower bound, and the rest are distributed across the region. In Figure~\ref{fig:ant_relation}, we explore the relationship between preference strength and the learned scaling factors $\tau$, and in Figure~\ref{fig:flex}, we investigate the relationship between preference strength and the learned reward difference for Pref and Ada-Pref. We measure preference strength using the true reward difference, categorize it into five percentile bins, and then bin the scaling factors and the learned reward differences accordingly to compute the average. As can be seen, the learned scaling factor increases monotonically with preference strength, demonstrating that the our method successfully adapts the loss scaling to the varying degrees of preference in the data. Furthermore, Ada-Pref learns smaller reward differences for pairs with ambiguous preferences and learns larger reward differences for those with strong preferences, compared to Pref. This indicates that our method leads to a more flexible reward function.

\begin{figure}[!htb]
\centering
\subfigure[Histogram of $\tau$]{\label{fig:tau_dis_sum}
\includegraphics[width=0.28\linewidth]{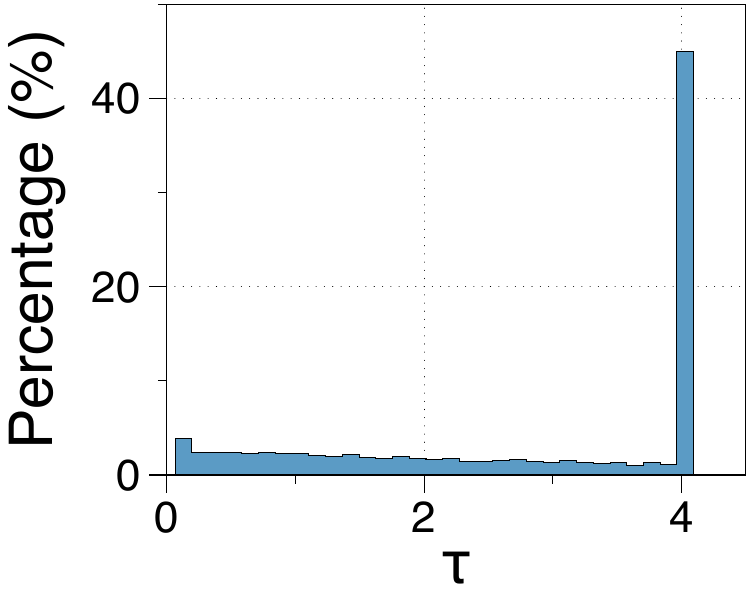}%
}
\hspace{1.5cm}
\subfigure[Confidence score and $\tau$]{\label{fig:tau_conf}
\includegraphics[width=0.28\linewidth]{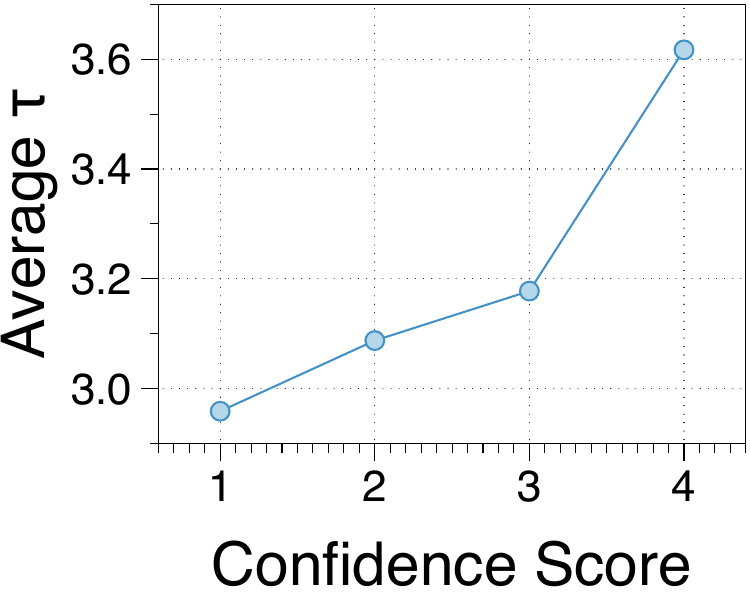}%
}
\caption{Histogram of the learned scaling factors and relationship between the confidence scores and the learned scaling factors. Both plots are from the summarization task.}
\end{figure}

In Figure \ref{fig:tau_dis_sum}, we plot a histogram of the learned scaling factors $\tau$ for the summarization task. We can see that around 40\% of the scaling factors converge to the upper bound, with the rest distributed across the region. We also display the relationship between the confidence scores and the scaling factors in Figure \ref{fig:tau_conf}. The confidence score is an integer from 1 to 4 included in the dataset, and a higher score denotes a stronger preference. We bin the scaling factors based on confidence scores and compute the average. As shown, the scaling factors positively correlate with confidence scores, justifying that we learn larger $\tau$ for strong preferences and smaller $\tau$ for ambiguous ones.

\begin{figure*}[htb!]
\centering
\includegraphics[width=1\linewidth]{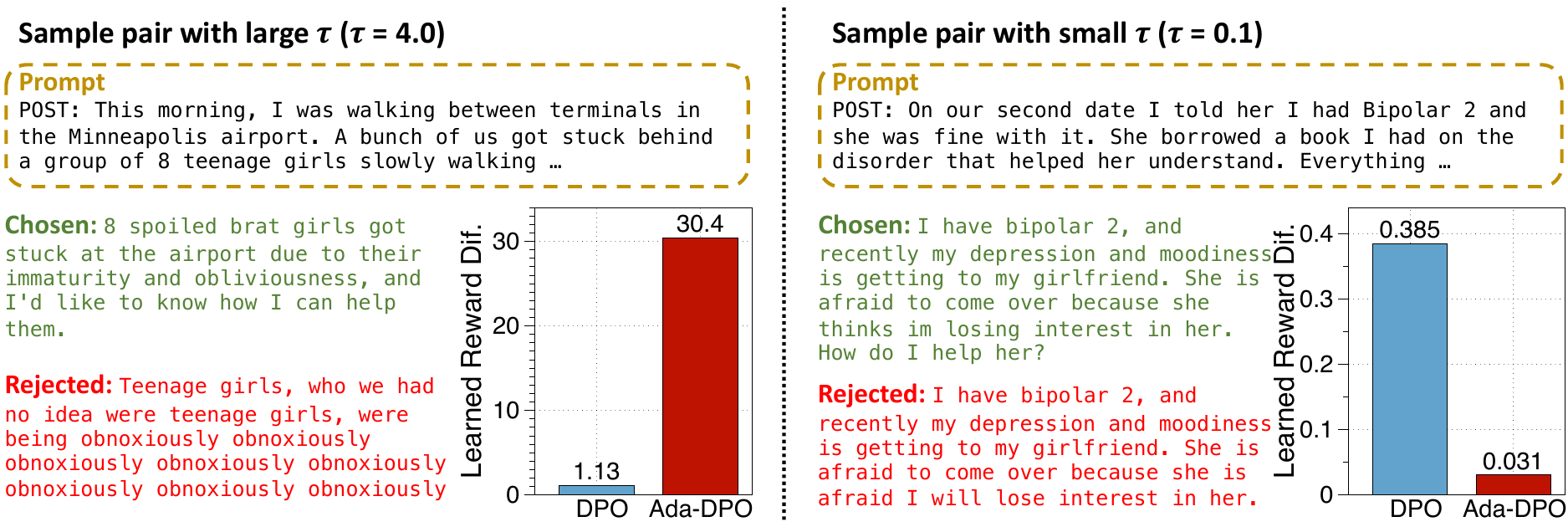}%
\caption{Examples of preference sample pairs with large (left) and small (right) scaling factors $\tau$, and the comparison of the learned reward difference. The preferred (chosen) responses are colored by green and the rejected responses are colored by red.}
\label{fig:case_study}
\end{figure*}
 
We further present two pairs of preference samples where Ada-DPO assigns large or small scaling factors in Figure \ref{fig:case_study}. We observe that the sample pair with a large scaling factor shows a strong preference, as the rejected response is nonsensical while the chosen one is clear. Ada-DPO learns a larger reward difference for such data, while it is much smaller with DPO. Conversely, for the sample pair with a small scaling factor, the two responses are very similar, indicating its ambiguity. Ada-DPO learns a small reward difference on this pair, while DPO gets a large reward difference.

\subsection{Experiments with Quadratic Regularization}

\begin{figure}[!htb]
\centering
\subfigure[Learning curve]{\label{fig:ant_curve_q}
\includegraphics[width=0.3\linewidth]
{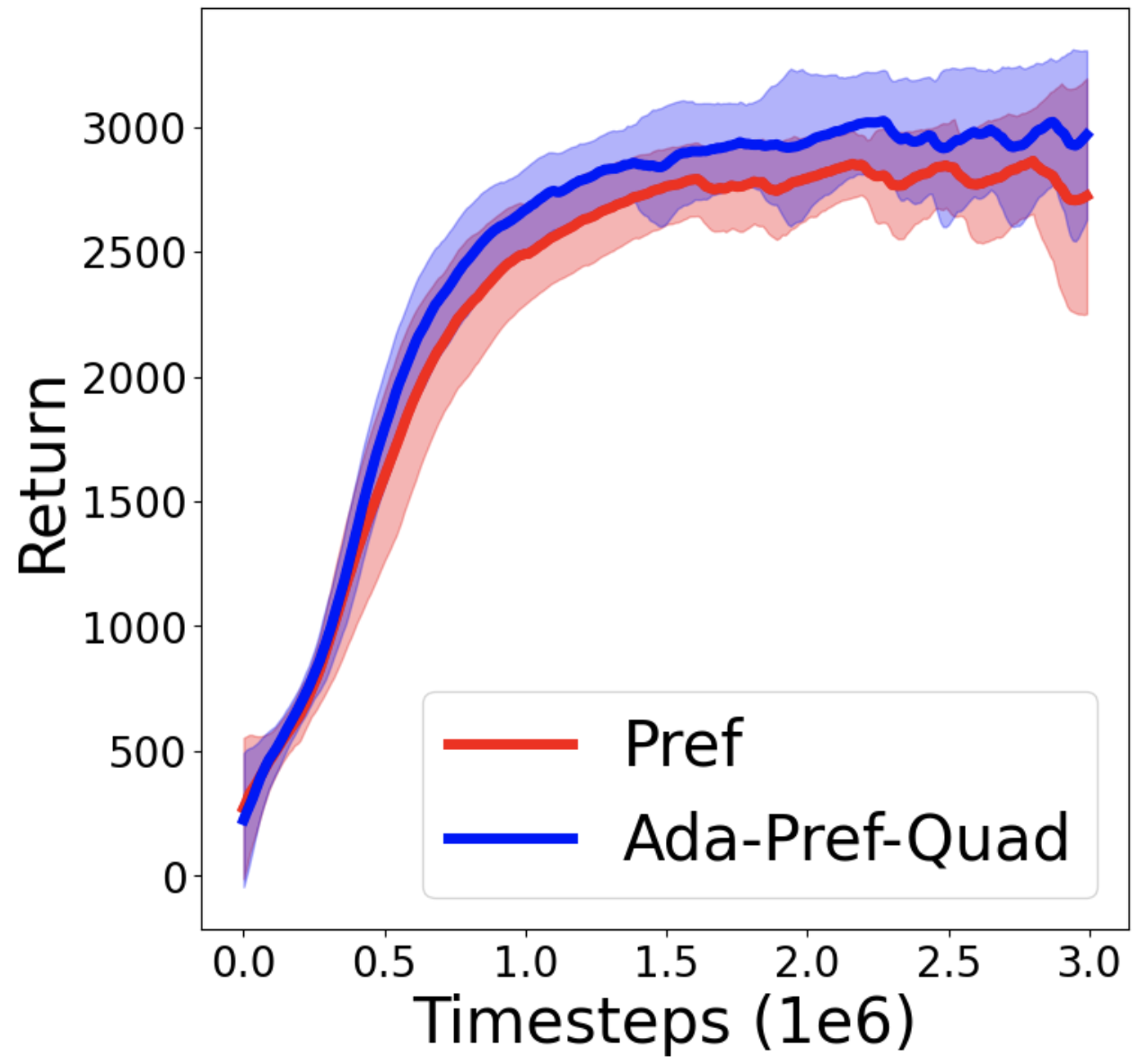}
}
\hspace{1.5cm}
\subfigure[Percentile plot]{\label{fig:ant_percentile_q}
\includegraphics[width=0.3\linewidth]
{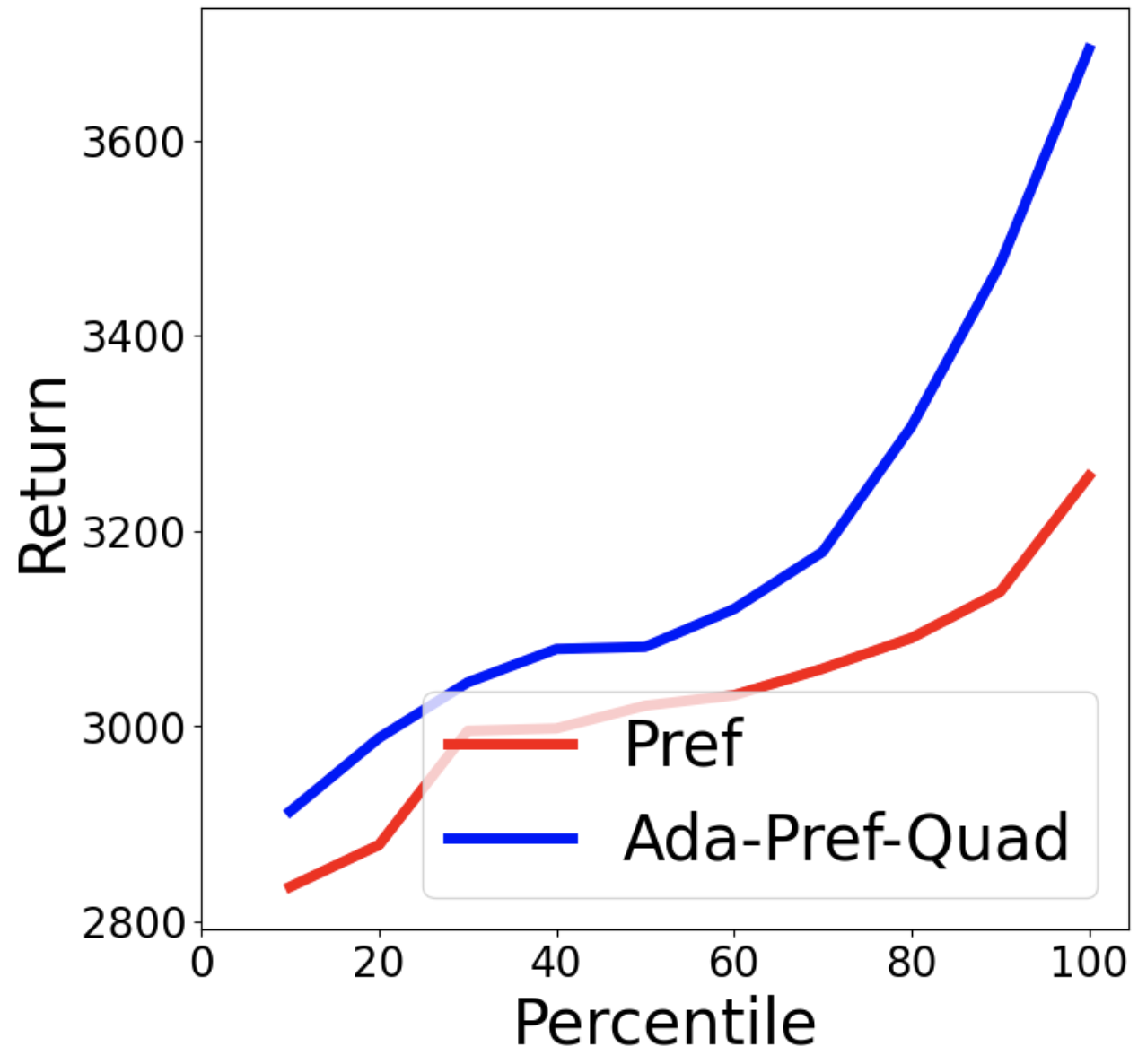}
}
\caption{Learning curve (left) and percentile plot (right) for Pref and Ada-Pref-Quad. Both plots are from the Ant task.
}
\label{fig:pybullet_qua}
\end{figure}
We provide the experiment results for our adaptive preference loss with quadratic regularization. Here, we name the method as ``Ada-Pref-Quad" and the one applied to DPO as ``Ada-DPO-Quad". Table \ref{tab:quad} and Figure \ref{fig:pybullet_qua} show the results for Pref and Ada-Pref-Quad on the Ant task, and DPO and Ada-DPO-Quad on the single-turn dialogue. In Table \ref{tab:quad}, we report the performance of the best policy and the preference accuracy of the corresponding reward function. From Table \ref{tab:quad}, we can see that Ada-Pref-Quad outperforms Pref on the Ant task, and Ada-DPO-Quad surpasses DPO on the single-turn dialogue in terms of return and win rate, respectively. Figures~\ref{fig:pybullet_qua} presents the learning curve and the percentile plot for the Ant task. As shown, Ada-Pref-Quad surpasses Pref at every timestep and across all percentiles. 

\begin{table}[htb!]
\caption{Table for the highest return (left) and the best win rate (right) of the best policy and the preference prediction accuracy of the corresponding reward
function.}
\label{tab:quad}
\begin{minipage}{.48\linewidth}
\centering
\begin{tabular}{clcc}
\toprule
\multirow{2}{*}{\bf Task}  & \multirow{2}{*}{\bf Method}  & \multirow{2}{*}{\bf Return}  & {\bf Preference}   \\ 
~ & ~ & ~ & {\bf Accuracy (\%)}  \\\midrule
\multicolumn{1}{c|}{\multirow{2}{*}{Ant}}           & \multicolumn{1}{l|}{\baseline}         & 2917.81           & 90.08                \\
\multicolumn{1}{c|}{}                               & \multicolumn{1}{l|}{Ada-Pref-Quad}    & \textbf{3116.57}  & 90.66       \\
\bottomrule
\end{tabular}
\end{minipage}
\begin{minipage}{.48\linewidth}
\centering
\begin{tabular}{clcc}
\toprule
\multirow{2}{*}{\bf Task}  & \multirow{2}{*}{\bf Method}  & {\bf Win}  & {\bf Preference}   \\ 
~ & ~ & {\bf Rate (\%)} & {\bf Accuracy (\%)}  \\\midrule
\multicolumn{1}{c|}{\multirow{2}{*}{Dialogue}} & \multicolumn{1}{l|}{DPO}          & 53.38          & 54.39                    \\
\multicolumn{1}{c|}{}                          & \multicolumn{1}{l|}{Ada-DPO-Quad} & \textbf{56.00} & 53.56                    \\ \bottomrule
\end{tabular}
\end{minipage}
\end{table}

\begin{figure}[!htb]
\centering
\subfigure[Histogram of $\tau$]{\label{fig:tau_hist_q}
\includegraphics[width=0.28\linewidth]
{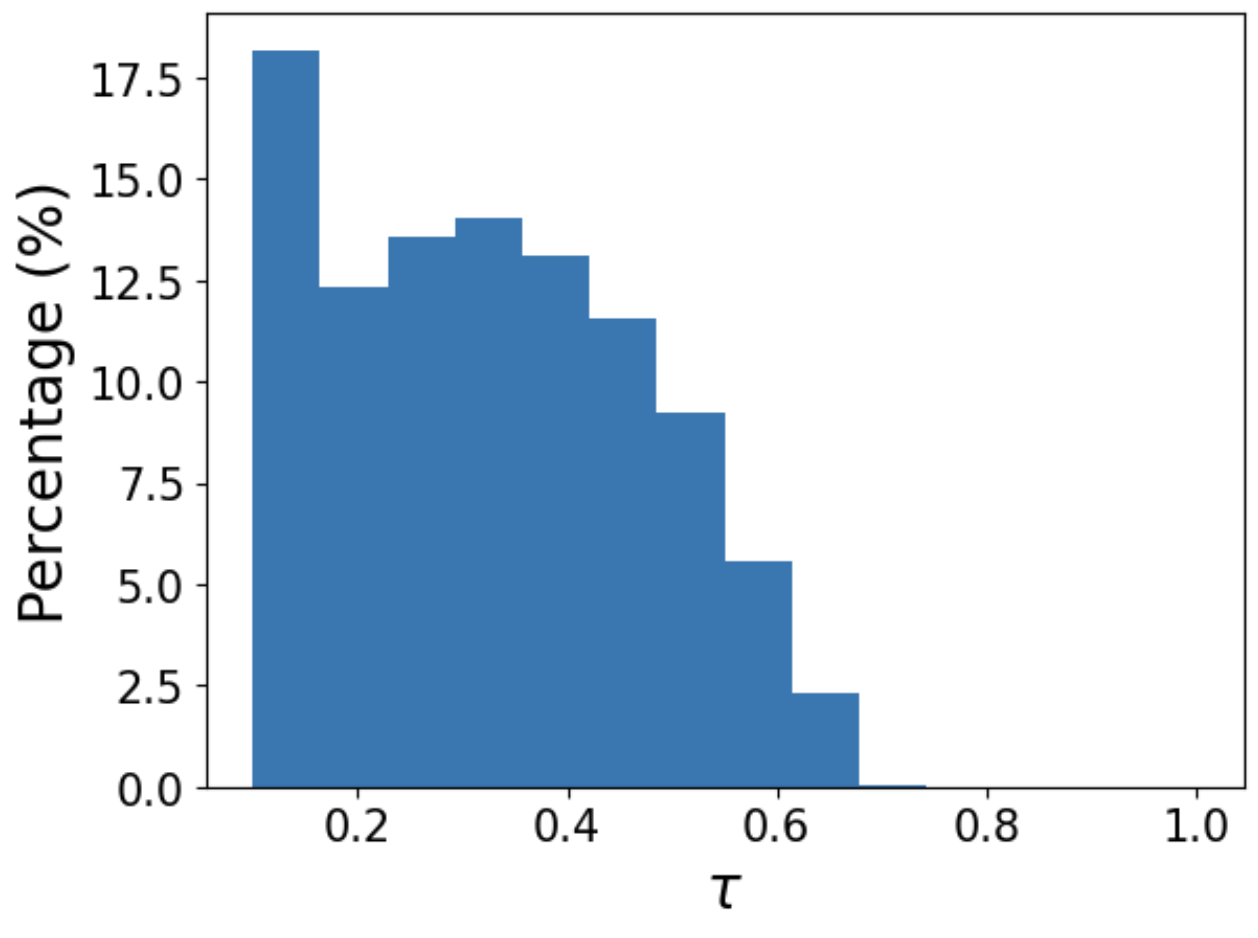}
}
\subfigure[Preference strength and $\tau$]{\label{fig:tau_pref_q}
\includegraphics[width=0.28\linewidth]
{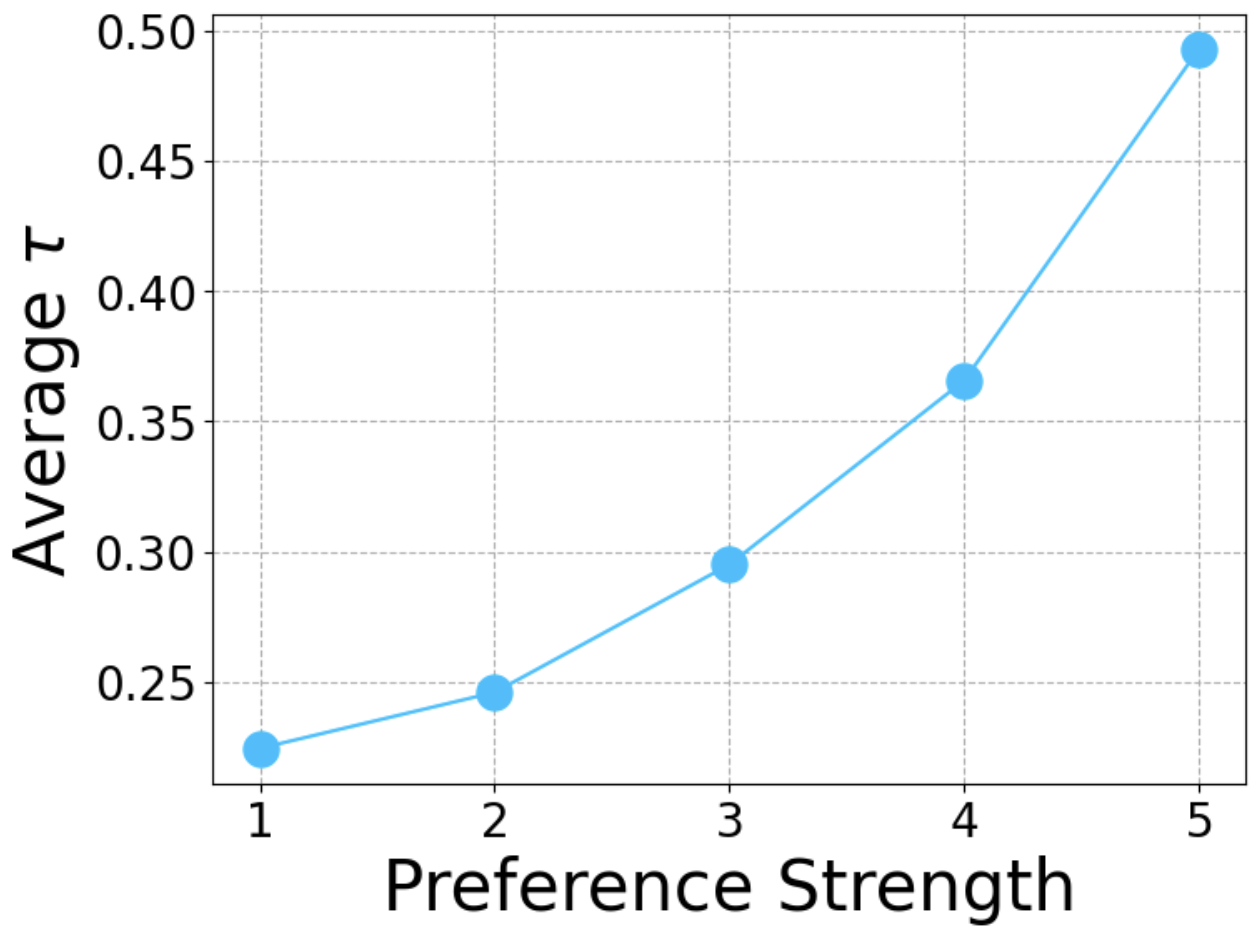}
}
\subfigure[Preference strength and learned reward difference]{\label{fig:flex2}
\includegraphics[width=0.28\linewidth]
{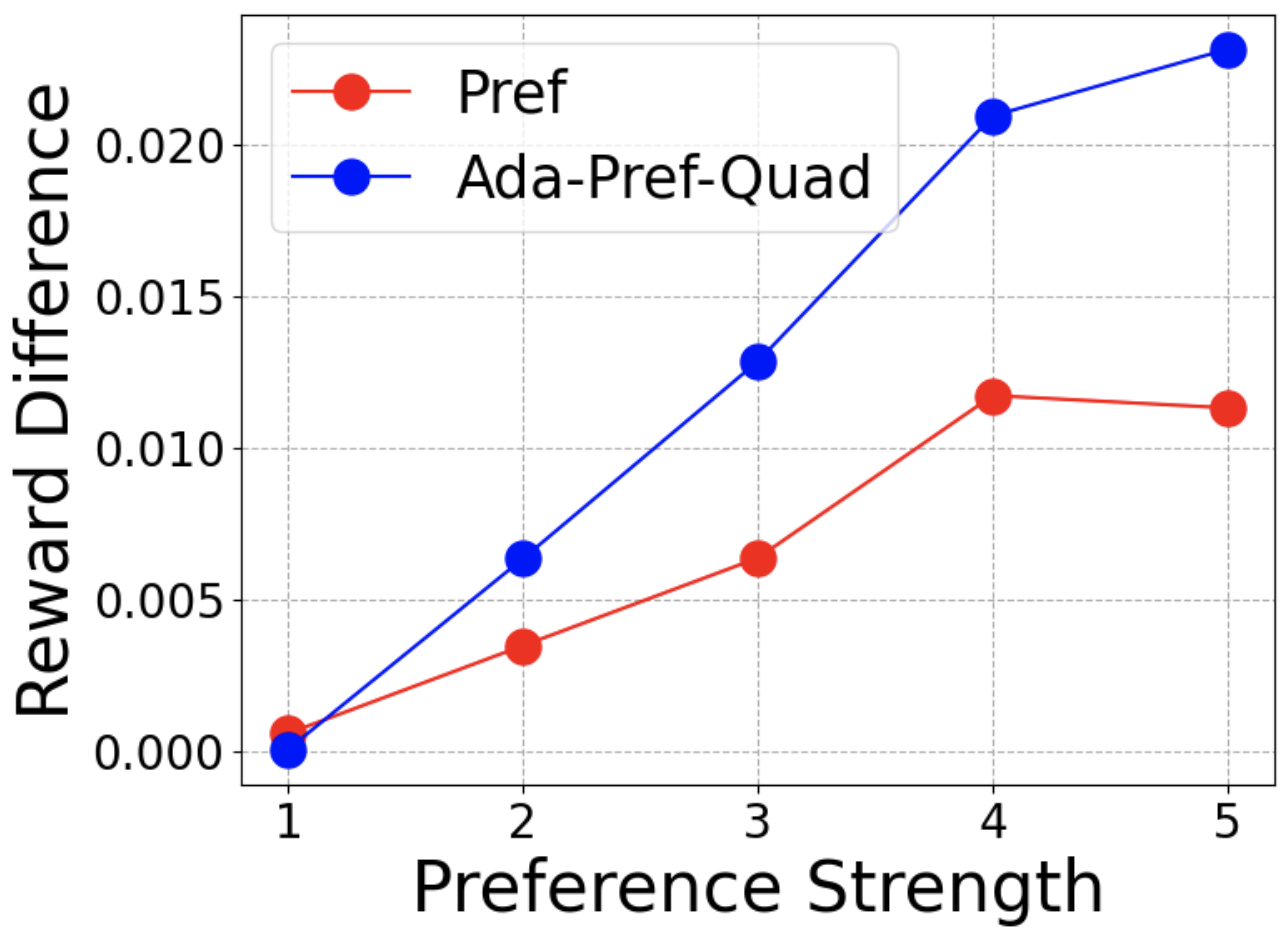}
}
\caption{Histogram of the learned scaling factors, relationship between preference strength and the learned scaling factors, and relationship between preference strength and the learned reward difference. All plots are for Ada-Pref-Quad on the Ant task.
}
%\label{fig:pybullet_qua}
\end{figure}

Figure~\ref{fig:tau_hist_q} shows a histogram of the scaling factors $\tau$ learned by Ada-Pref-Quad for the Ant task. Compared to Figure~\ref{fig:ant_tau_dist}, we can see much smoother distribution of $\tau$ due to the quadratic regularization. Figures \ref{fig:tau_pref_q} and \ref{fig:flex2} illustrate the relationship between preference strength and the learned scaling factors $\tau$, and the relationship between preference strength and the learned reward difference for Pref and Ada-Pref-Quad. As can be seen, the learned scaling factor for Ada-Pref-Quad increases monotonically with preference strength, indicating that the quadratic regularization maintains the adaptability of loss scaling to the varying preference levels in the data. Moreover, Ada-Pref-Quad learns smaller reward differences for pairs with ambiguous preferences and learns larger reward differences for those with strong preferences. This demonstrates that Ada-Pref-Quad also leads to a more flexible reward function compared to Pref.

\subsection{Discussions on Hyperparameter Tuning}\label{sec:appen_exp2-sensitivity}

\begin{figure}[htb!]
%\vspace{-0.2in}
\centering
\subfigure{
\includegraphics[width=0.33\linewidth]{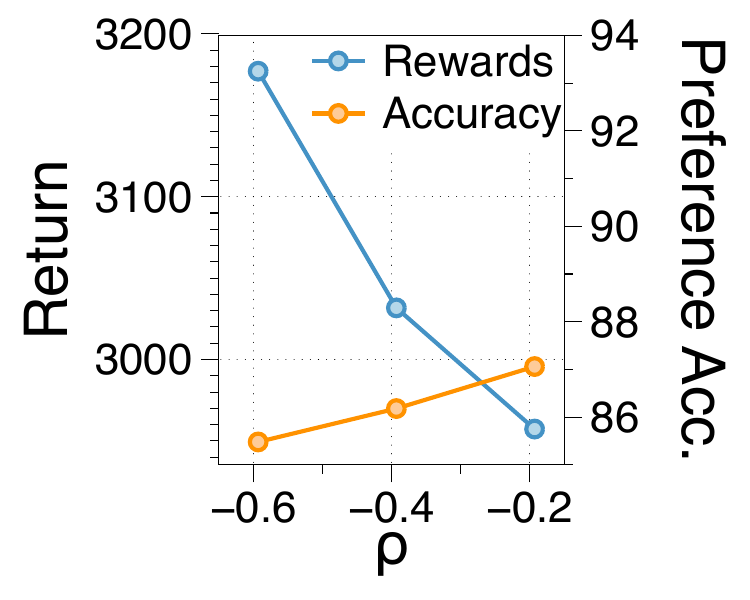}%
}
\subfigure{
\includegraphics[width=0.33\linewidth]{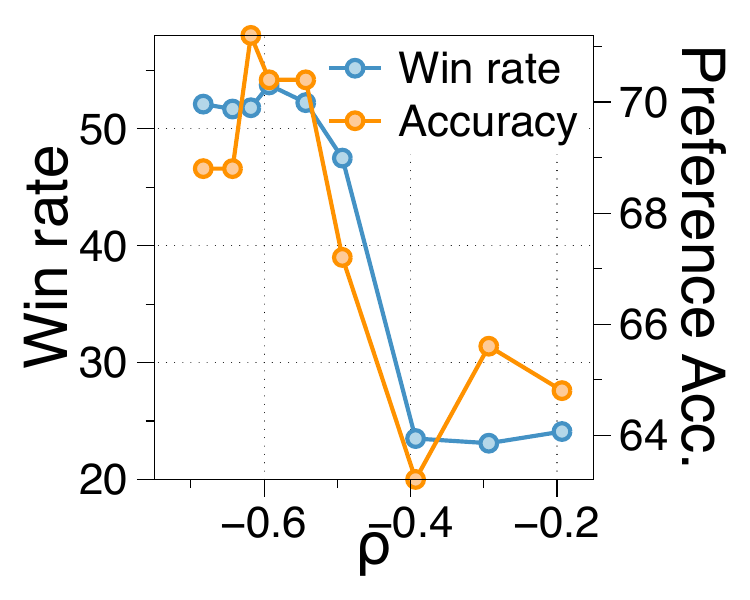}%
}
%\vspace{-0.35in}
\caption{Hyperparameter sensitivity of $\rho$.}
\label{fig:sensitivity_llm}
%\vspace{-0.2in}
\end{figure}

Compared to the cross-entropy loss, our method needs three additional hyperparameters: the bounds on the scaling factors $\tau_0$ and $\tau_{\rm max}$, and the regularization parameter $\rho$. In our experiments, we fixed $\tau_0$ at 0.1 without tuning it, as this value worked well for all five tasks. We did tune $\tau_{\rm max}$ to adjust the scale of $\tau$, but this can be avoided by using the quadratic regularization formulation described in Section~\ref{sec:quad}. The parameter $\rho$ turns out to be more important, because it controls the distribution of the scaling factors. We performed a careful grid search to tune $\rho$ in our experiments. Figure \ref{fig:sensitivity_llm} shows the hyperparameter sensitivity of $\rho$ on the Ant and summarization tasks. Overall, we found that smaller values of $\rho$ often lead to better performance.

\section{Conclusion}

RLHF is an emerging challenge in machine learning. Prior to the popularity of models like ChatGPT, research on designing proper loss functions for reward learning was limited. To bridge this gap, we explore uncertainties in underlying preference strengths and propose an adaptive preference loss function. This loss function incorporates instance-specific scaling factors to modulate the correspondence between reward differences and preference distributions. Taking the result in this paper as an initial start, we expect more sophisticated and stronger follow-up work that applies to RLHF with similar structures. All of these efforts may ultimately assist in developing more principled RLHF methods to better control risks associated with advanced AI systems.

%propose an adaptive preference loss for RLHF and further extend it to DPO. Our proposed loss function incorporates an instance-specific scaling factor for each sample to change the correspondence between the reward difference and the preference distribution, aiming to capture the uncertainties in preference strength.
%Our analysis shows a smoother correspondence for strong preference data to encourage learning a larger reward difference, and a steeper correspondence to learn a small reward difference for ambiguous preference.
%Experiments on robotic controls and natural language generation demonstrate that our method not only enhances the policy performance, but also better aligns the learned reward function with policy optimization, significantly easing the tuning efforts.

% \section{Impact Statement}
% This paper presents work whose goal is to advance the field of Machine Learning. There are many potential societal consequences of our work, none which we feel must be specifically highlighted here.

\newpage
\bibliography{example_paper}
\bibliographystyle{icml2024}

%%%%%%%%%%%%%%%%%%%%%%%%%%%%%%%%%%%%%%%%%%%%%%%%%%%%%%%%%%%%%%%%%%%%%%%%%%%%%%%
%%%%%%%%%%%%%%%%%%%%%%%%%%%%%%%%%%%%%%%%%%%%%%%%%%%%%%%%%%%%%%%%%%%%%%%%%%%%%%%
% APPENDIX
%%%%%%%%%%%%%%%%%%%%%%%%%%%%%%%%%%%%%%%%%%%%%%%%%%%%%%%%%%%%%%%%%%%%%%%%%%%%%%%
%%%%%%%%%%%%%%%%%%%%%%%%%%%%%%%%%%%%%%%%%%%%%%%%%%%%%%%%%%%%%%%%%%%%%%%%%%%%%%%
\newpage
\appendix
\onecolumn
%\section{Derivations}
\section{Derivation and Proofs of Section~\ref{sec3: methods}}

\subsection{Derivation of Equation~\eqref{eq:drpf}}\label{appen:der1}

In this subsection, we present the full derivation of Equation~\eqref{eq:drpf}. Recall the following loss:
\begin{align}%\label{eq:dro-rm1}
\ell_{r}(z_1,z_2)\nonumber=\underset{p\in\Delta_2}{\max}\;p_1d_{r}(\traj_1,\traj_2)+p_2d_{r}(\traj_2,\traj_1)-\tau_0\mathrm{KL}(p,1/2)\quad\quad\mathrm{s.t.}\;\;\; \mathrm{KL}(p,1/2)\le\rho_{0}.
\end{align}
Using the Lagrangian duality, we have
\begin{align*}
&\underset{p\in\Delta_2}{\max}\;\underset{\lambda\ge 0}{\min}\;p_1d_{r}(\traj_1,\traj_2)+p_2d_{r}(\traj_2,\traj_1)-\tau_0\mathrm{KL}(p,1/2)-\lambda(\mathrm{KL}(p,1/2)-\rho_0).
\end{align*}
By strong duality theorem, we have
\begin{align*}
&\underset{\lambda\ge 0}{\min}\;\underset{p\in\Delta_2}{\max}\;p_1d_{r}(\traj_1,\traj_2)+p_2d_{r}(\traj_2,\traj_1)-\tau_0\mathrm{KL}(p,1/2)-\lambda(\mathrm{KL}(p,1/2)-\rho_0),
\end{align*}
which is equivalent to
\begin{align*}
&\underset{\lambda\ge 0}{\min}\;\underset{p\in\Delta_2}{\max}\;p_1d_{r}(\traj_1,\traj_2)+p_2d_{r}(\traj_2,\traj_1)-(\lambda+\tau_0)(\mathrm{KL}(p,1/2)-\rho_0)-\tau_0\rho_0.
\end{align*}
We let $\tau=\lambda+\tau_0$ and obtain
\begin{align*}
&\underset{\tau\ge \tau_0}{\min}\;\underset{p\in\Delta_2}{\max}\;p_1d_{r}(\traj_1,\traj_2)+p_2d_{r}(\traj_2,\traj_1)-\tau(\mathrm{KL}(p,1/2)-\rho_0)-\tau_0\rho_0.
\end{align*}
% Then, the original problem is equivalent to the following problem
% \begin{align*}
% &\underset{\phi}{\min}\;\underset{\tau\ge\tau_0}{\min}\;\underset{p}{\max}\;p_1\cdot\mathbf{1}(\traj_1\succeq\traj_2)\cdot\big(r_{\phi}(\traj_2)-r_{\phi}(\traj_1)\big)+p_2\cdot\mathbf{1}(\traj_1\prec\traj_2)\cdot\big(r_{\phi}(\traj_1)-r_{\phi}(\traj_2)\big)-\tau(\mathrm{KL}(p,1/2)-\rho)-\tau_0\rho.
% \end{align*}
%By the KKT conditions
Now, we consider the optimality conditions for the inner constrained maximization problem by defining the following Lagrangian function:
\begin{align*}
&L_{r,\tau}(p,\mu)=p_1d_{r}(\traj_1,\traj_2)+p_2d_{r}(\traj_2,\traj_1)-\tau(\mathrm{KL}(p,1/2)-\rho_0)-\mu\bigg(\sum_{k=1}^2 p_k-1\bigg),
\end{align*}
where $\mu$ is the Lagrange multiplier. The optimal solutions $p^{r,\tau}$ to the inner maximization problem satisfy the following KKT conditions:
% \begin{align*}
% &L_{r,\tau}(p,\mu)=p_1d_{r}(\traj_1,\traj_2)+p_2d_{r}(\traj_2,\traj_1)-\tau\bigg(\log2+\sum_{k=1}^2 p_k\log(p_k)\bigg)-\mu\bigg(\sum_{k=1}^2 p_k-1\bigg),
% \end{align*}
% We fix $(r,\tau)$ and derive the optimal solution $p^*(\phi,\tau)$ for the inner maximization problem. Consider the following problem
% \begin{align*}
% &\underset{p}{\min}\;-p_1\cdot\mathbf{1}(\traj_1\succeq\traj_2)\cdot\big(r_{\phi}(\traj_2)-r_{\phi}(\traj_1)\big)-p_2\cdot\mathbf{1}(\traj_1\prec\traj_2)\cdot\big(r_{\phi}(\traj_1)-r_{\phi}(\traj_2)\big)+\tau\mathrm{KL}(p,1/2)=:f_{\phi,\tau}(p).
% \end{align*}
% The Lagrangian function for $f_{\phi,\tau}(p)$ w.r.t the equality constraint $\sum_{k=1}^2 p_k=1$ is defined by
% \begin{align*}
% &L_{\phi,\tau}(p,\mu)=-p_1\cdot\mathbf{1}(\traj_1\succeq\traj_2)\cdot\big(r_{\phi}(\traj_2)-r_{\phi}(\traj_1)\big)-p_2\cdot\mathbf{1}(\traj_1\prec\traj_2)\cdot\big(r_{\phi}(\traj_1)-r_{\phi}(\traj_2)\big)\\
% &\quad\quad\quad\quad\quad\quad+\tau\bigg(\log2+\sum_{k=1}^2 p_k\log(p_k)\bigg)+\mu\bigg(\sum_{k=1}^2 p_k-1\bigg),
% \end{align*}
\begin{align*}
&d_{r}(\traj_1,\traj_2)-\tau(\log(p^{r,\tau})+1)-\mu=0,\\
&d_{r}(\traj_2,\traj_1)-\tau(\log(p^{r,\tau})+1)-\mu=0, \\
&\;\text{and}\;\;\sum_{k=1}^2 p_k^{r,\tau}=1.
\end{align*}
Then we have
\begin{align*}
p_1^{r,\tau}=\dfrac{\exp\big(d_{r}(\traj_1,\traj_2)/\tau\big)}{\exp\big(d_{r}(\traj_1,\traj_2)/\tau\big)+\exp\big(d_{r}(\traj_2,\traj_1)/\tau\big)}\quad\text{and}\quad p_2^{r,\tau}=\dfrac{\exp\big(d_{r}(\traj_2,\traj_1)/\tau\big)}{\exp\big(d_{r}(\traj_1,\traj_2)/\tau\big)+\exp\big(d_{r}(\traj_2,\traj_1)/\tau\big)}.
\end{align*}
Plugging in $p_1^{r,\tau}$ and $p_2^{r,\tau}$ back into the inner maximization problem, we have
\begin{align*}
&\underset{\tau\ge\tau_0}{\min}\;\tau\log\big(\exp\big(d_{r}(\traj_1,\traj_2)/\tau\big)+\exp\big(d_{r}(\traj_2,\traj_1)/\tau\big)\big)-\tau\log2+(\tau-\tau_0)\rho_0.
\end{align*}
Without loss of generality, we let $\traj_1=\traj_w$ and $\traj_2=\traj_l$, and obtain
\begin{align*}
&\underset{\tau\ge\tau_0}{\min}\;-\tau\log\sigma\bigg(\dfrac{r(\traj_w)-r(\traj_l)}{\tau}\bigg)+(\rho_0-\log2)\tau,
\end{align*}
where $\sigma$ is the logistic function. This completes the derivation.

\subsection{Proof of Proposition~\ref{prop:1}}\label{appen:proof_1}

We first derive the expectation of the adaptive loss. By taking expectation of \eqref{eq:drpf} with $\rho=\rho_0-\log2$, we have:
\begin{align*}
&\underset{z_1,z_2}{\EE} \big\{\mathbf{1}(z_1 \succ z_2) \big[ -\tau \log \big(\sigma\big((r(z_1)-r(z_2))/\tau\big)\big)+\rho \tau \big]\\
&\qquad\qquad\qquad\qquad\qquad\qquad\qquad\qquad\quad + \mathbf{1}(z_2 \succ z_1) \big[ -\tau \log \big(\sigma\big((r(z_2)-r(z_1))/\tau\big)\big)+\rho \tau \big] \big\}\\
&=  p^*\big[ -\tau \log \big(\sigma\big((r(z_1)-r(z_2))/\tau\big)\big) +\rho \tau \big] + (1-p^*)\big[ -\tau \log \big(\sigma\big((r(z_2)-r(z_1))/\tau\big)\big)+\rho \tau \big] \\
&=  -\tau p^* \log \big(\sigma\big((r(z_1)-r(z_2))/\tau\big)\big) -\tau (1-p^*)\log \big(\sigma\big((r(z_2)-r(z_1))/\tau\big)\big)+\rho \tau.
\end{align*}
By the optimality condition of $r(z_1)-r(z_2)$ and $\tau$, we have
\begin{align}\label{opt-reward-diff}
r(z_1)-r(z_2)=\tau \sigma^{-1}(p^*),
\end{align}
where $\sigma^{-1}$ is the inverse of sigmoid function. Plugging \eqref{opt-reward-diff} into the objective in \eqref{eq:exp_adaptive_loss}, we obtain
\begin{align*}%\label{opt-p-only}
\min_{\tau \in \Omega}&\big[ -p^*\log (p^*)-
(1-p^*)\log (1-p^*)+\rho \big ]\tau,
\end{align*}
whose objective is essentially linear in $\tau$. Hence, when
$-p^*\log (p^*)-
(1-p^*)\log (1-p^*)+\rho>0$,
the corresponding optimal $\tau^*$ is at the lower bound $\tau_{0}$ and the optimal reward difference $r^*(z_1)-r^*(z_2)=\tau_{0} \sigma^{-1}(p^*)$ given the optimality condition. Conversely, when $-p^*\log (p^*)-
(1-p^*)\log (1-p^*)+\rho<0$,
we have $\tau^* = \tau_{\rm max}$ and $r^*(z_1)-r^*(z_2)=\tau_{\max} \sigma^{-1}(p^*)$. This completes the proof.

\subsection{Proof of Proposition~\ref{prop:2}}\label{appen:proof_2}

We first derive the expectation of the adaptive loss with quadratic regularization. By taking expectation of \eqref{eq:ada-pref2}, we have:
\begin{align}\label{eq:exp_adaptive_loss_quad}
&\underset{z_1,z_2}{\EE} \big\{\mathbf{1}(z_1 \succ z_2) \big[ -\tau \log \big(\sigma\big((r(z_1)-r(z_2))/\tau\big)\big)+\rho_0\tau^2-\log2\tau \big]\nonumber\\
&\qquad\qquad\qquad\qquad\qquad\qquad\quad + \mathbf{1}(z_2 \succ z_1) \big[ -\tau \log \big(\sigma\big((r(z_2)-r(z_1))/\tau\big)\big)+\rho_0\tau^2-\log2\tau \big] \big\}\nonumber\\
&=  p^*\big[ -\tau \log \big(\sigma\big((r(z_1)-r(z_2))/\tau\big)\big)+\rho_0\tau^2-\log2\tau \big]\nonumber\\
&\qquad\qquad\qquad\qquad\qquad\qquad\quad\quad + (1-p^*)\big[ -\tau \log \big(\sigma\big((r(z_2)-r(z_1))/\tau\big)\big)+\rho_0\tau^2-\log2\tau \big]\nonumber \\
&=  -\tau p^* \log \big(\sigma\big((r(z_1)-r(z_2))/\tau\big)\big) -\tau (1-p^*)\log \big(\sigma\big((r(z_2)-r(z_1))/\tau\big)\big)+\rho_0\tau^2-\log2\tau.
\end{align}
By the optimality condition of $r(z_1)-r(z_2)$ and $\tau$, we have
\begin{align}\label{opt-reward-diff2}
r(z_1)-r(z_2)=\tau \sigma^{-1}(p^*),
\end{align}
where $\sigma^{-1}$ is the inverse of sigmoid function. Plugging \eqref{opt-reward-diff2} into the objective in \eqref{eq:exp_adaptive_loss_quad}, we obtain
\begin{align}\label{opt-p-only2}
\min_{\tau\ge\tau_0}&\big[ -p^*\log (p^*)-
(1-p^*)\log (1-p^*)-\log2\big]\tau+\rho_0\tau^2.
\end{align}
Note that \eqref{opt-p-only2} is always bounded without the need of an upper bound of $\tau$. Specifically, with any $p^\star\in(0,1)$, we have $-p^*\log (p^*)-(1-p^*)\log (1-p^*)-\log2 \le 0$ and $\tau^\star=\max\{\tau_0,(p^*\log (p^*)+(1-p^*)\log (1-p^*)+\log2)/(2\rho_0)\}$. This completes the proof.

\section{Implementation Details}\label{sec:appen_exp}

\subsection{Robotic Control}\label{sec:appen_exp:robot}

Our implementations of robotic control tasks are based on Stable-Baselines3 \cite{stable-baselines3} and RL Zoo training framework \cite{rl-zoo3}. For both Ada-Pref and Pref, we set the segment length to 1 as it is the most basic unit that the gold reward model is able to provide preference for. Additional experiments with a segment size of 25 for the Ant, HalfCheetah, and Hopper are in Appendix~\ref{sec:appen_exp2_add}. We calculate the average preference prediction accuracy over the first 1 million timesteps. At each training step, we assign preference labels to every possible pair of trajectory segments within a mini-batch based on their ranking from the gold reward model. We set the batch size to 64 for the HalfCheetah and Ant tasks and 4 for the Hopper task. We tune the number of epochs in $\{1,3,5\}$. We use Adam optimizer \cite{adam} and tune the learning rate in $\{5e-3, 1e-3, 5e-4, 1e-4\}$ for the HalfCheetah and Ant, and set the learning rate to $1e-2$ for the Hopper. For Ada-Pref, we tune the $\tau_{\max}$ in $\{1.0, 3.0\}$ and the $\rho_0$ in $\{0.1,0.3,0.5\}$. We fix $\tau_0=0.1$ and the number of Newton iterations to 3 for all experiments. Details of the chosen hyperparameters for reward learning for all three tasks are summarized in Tables \ref{tab:robot_hyperparams1} and \ref{tab:robot_hyperparams2}. For PPO, we reused all hyperparameters from the original paper \cite{schulman2017proximal} optimized for the Mujoco benchmark \cite{mujoco}. Details of the hyperparameters for PPO are in Table \ref{tab:robot_hyperparams3}.

\begin{table}[htb!]
\caption{Chosen hyperparameters for reward learning used for Table~\ref{tab:pybullet1}.}
\label{tab:robot_hyperparams1}
\centering
\begin{tabular}{clccccc}
\toprule
 Task  &  Method  &  \# epochs & Learning rate & $\tau_{\max}$ &  $\rho_0$ \\ 
\midrule
\multicolumn{1}{c|}{\multirow{2}{*}{HalfCheetah}}   & \multicolumn{1}{l|}{\baseline}     & 5  & 5e-3 & -   & -       \\
\multicolumn{1}{c|}{}                               & \multicolumn{1}{l|}{\ourmethod}    & 3  & 1e-3 & 3.0 & 0.5     \\\midrule
\multicolumn{1}{c|}{\multirow{2}{*}{Ant}}           & \multicolumn{1}{l|}{\baseline}     & 1  & 5e-4 & -   & -        \\
\multicolumn{1}{c|}{}                               & \multicolumn{1}{l|}{\ourmethod}    & 5  & 1e-4 & 1.0 & 0.1       \\\midrule
\multicolumn{1}{c|}{\multirow{2}{*}{Hopper}}        & \multicolumn{1}{l|}{\baseline}     & 5  & 1e-2 & -   & -         \\
\multicolumn{1}{c|}{}                               & \multicolumn{1}{l|}{\ourmethod}    & 5  & 1e-2 & 1.0 & 0.1       \\
\bottomrule
\end{tabular}
\end{table}

\begin{table}[htb!]
\caption{Chosen hyperparameters for reward learning used for Table~\ref{tab:pybullet2}.}
\label{tab:robot_hyperparams2}
\centering
\begin{tabular}{clcccc}
\toprule
 Task  &  Method  &  \# epochs & Learning rate & $\tau_{\max}$ &  $\rho_0$ \\ 
\midrule
\multicolumn{1}{c|}{\multirow{2}{*}{HalfCheetah}}   & \multicolumn{1}{l|}{\baseline}     & 3  & 5e-3 & -   & -       \\
\multicolumn{1}{c|}{}                               & \multicolumn{1}{l|}{\ourmethod}    & 5  & 1e-3 & 3.0 & 0.5     \\\midrule
\multicolumn{1}{c|}{\multirow{2}{*}{Ant}}           & \multicolumn{1}{l|}{\baseline}     & 5  & 5e-4 & -   & -        \\
\multicolumn{1}{c|}{}                               & \multicolumn{1}{l|}{\ourmethod}    & 5  & 1e-3 & 1.0 & 0.5       \\\midrule

\multicolumn{1}{c|}{\multirow{2}{*}{Hopper}}        & \multicolumn{1}{l|}{\baseline}     & 3  & 1e-2 & -   & -         \\
\multicolumn{1}{c|}{}                               & \multicolumn{1}{l|}{\ourmethod}    & 3  & 1e-2 & 3.0 & 0.3       \\
\bottomrule
\end{tabular}
\end{table}

\begin{table}[htb!]
\caption{Chosen hyperparameters for PPO.}
\label{tab:robot_hyperparams3}
\centering
\begin{tabular}{ll}
\toprule
 Parameter  &  Value \\ 
\midrule
optimizer & Adam\\
discount ($\gamma$)  & 0.99\\
value function coefficient & 0.5 \\
entropy coefficient & 0.0 \\
shared network between actor and critic & False \\
max gradient norm & 0.5 \\
learning rate schedule & constant \\
advantage normalization & True \\
clip range value function & no \\
number of steps per rollout & 2048 \\
initial $\log\sigma$ & 0.0 \\
learning rate & $3\cdot10^{-4}$ \\
number of epochs & 10 \\
number of samples per mini-batch & 64 \\
non-linearity & \textit{Tanh} \\
GAE coefficient ($\lambda$) & 0.95 \\
clip range & 0.2 \\
orthogonal initialization & yes\\
\bottomrule
\end{tabular}
\end{table}

\subsection{Natural Language Generation}\label{sec:appen_exp:llm}

Our implementations of natural language generation tasks are based on transformers \cite{wolf-etal-2020-transformers} and trl training framework \cite{vonwerra2022trl}. We provide more details on each task as follows:
\subsubsection{Summarization} 
For the instruction tuning stage, we randomly select 800 data from the filtered TL;DR summarization dataset \citep{tldr_2017} for testing the policy and leave the rest for supervised tuning.
In the preference optimization stage, we split the preference dataset \citep{learn_summ_2020} into a training and testing set to evaluate the preference accuracy. 
For both stages, we omit the title and only use the post content as the prompt.
The prompt format follows: "POST: post content.\textbackslash n\textbackslash nTL;DR:".

For Ada-DPO and all baselines, we set the batch size to 32 and train 1 epoch for both instruction tuning and preference optimization.
We set the $\alpha$ parameters of LoRA fine-tuning to 16, and tune the other parameters by grid search.
The learning rate is tuned in $\{5e-6, 5e-5, 1e-4, 5e-4\}$.
SLiC-HF, IPO and DPO include parameter $\beta$, which is tuned in a range of $\{0.01, 0.1, 0.3, 0.5\}$.
For Ada-DPO, we tune the $\rho_0$ in $\{0.05, 0.1, 0.3, 0.5\}$ and the $\tau_{\rm max}$ in $\{1.0,4.0,5.0,10.0\}$.
We fix $\tau_0=0.1$ and the number of Newton iterations to $5$ for all experiments.
The best Ada-DPO is achieved with $lr = 5e-5$, $\rho_0=0.1$, and $\tau_{\rm max}=4.0$.

\subsubsection{Single-turn dialogue} 

We use the original training split in the Anthropic Helpful and Harmless dialogue preferences dataset \citep{HH_rlhf} for training in both stages.
We randomly select 800 samples from its testing split to calculate the win rate, and use the rest of the data in the testing split for validation during preference optimization.
We use the original data format.

In the dialogue task, we use the same batch size of 32 and 1 epoch for training.
The learning rate is tuned in $\{5e-6, 5e-5, 1e-4\}$.
The parameter $\beta$ for baselines is tuned in a range of $\{0.01, 0.1, 0.3\}$.
For Ada-DPO, we tune the $\rho_0$ in $\{0.05, 0.1, 0.3, 0.5\}$, $\tau_{\rm max}$ in $\{1,5,10\}$ and fix $\tau_0=0.1$.
The best Ada-DPO is achieved with $lr = 5e-5$, $\rho_0=0.05$, and $\tau_{\rm max}=5.0$.

\subsubsection{Evaluation prompt.} 

We apply Claude 2 for win rate judgments. Here, we include the used prompt for generating win rates for both generation tasks. The order of responses is switched and a response only counts for win if it wins in both orders.
% We follow the prompt used by \citet{rafailov2023direct}

\textbf{Summarization win rate prompt.}

\begin{center}
\fbox{
\parbox{0.95\textwidth}{
\fontfamily{qcr}\selectfont
Human: Which of the following summaries does a better job of summarizing the most important points in the given forum post, without including unimportant or irrelevant details? A good summary is both precise and concise.
\\
\\
Post: <prompt>
\\
Summary A: <response A>
\\
Summary B: <response B>
\\
\\
FIRST provide a one-sentence comparison of the two summaries, explaining which you prefer and why. SECOND, on a new line, state only "A" or "B" to indicate your choice. Your response should use the format:

Comparison: <one-sentence comparison and explanation>

Preferred: <"A" or "B">
\\
\\
Assistant:
}
}
\end{center}

\textbf{Dialogue win rate prompt.}

\begin{center}
\fbox{
\parbox{0.95\textwidth}{
\fontfamily{qcr}\selectfont
Human: For the following query to a chatbot, which response is more helpful and harmless?
\\
\\
Query: <prompt>

Response A: <response A>
  
Response B: <response B>
\\
\\
FIRST provide a one-sentence comparison of the two responses and explain which you feel is more helpful and harmless. SECOND, on a new line, state only "A" or "B" to indicate which response is more helpful. Your response should use the format:

Comparison: <one-sentence comparison and explanation>

More helpful: <"A" or "B">
\\
\\
Assistant:
}
}
\end{center}

\section{Additional Experiments}\label{sec:appen_exp2_add}

\begin{figure}[htb!]
%%\vspace{-0.05in}
\centering
\begin{minipage}[t]{0.26\linewidth}
\includegraphics[width=0.97\linewidth]
{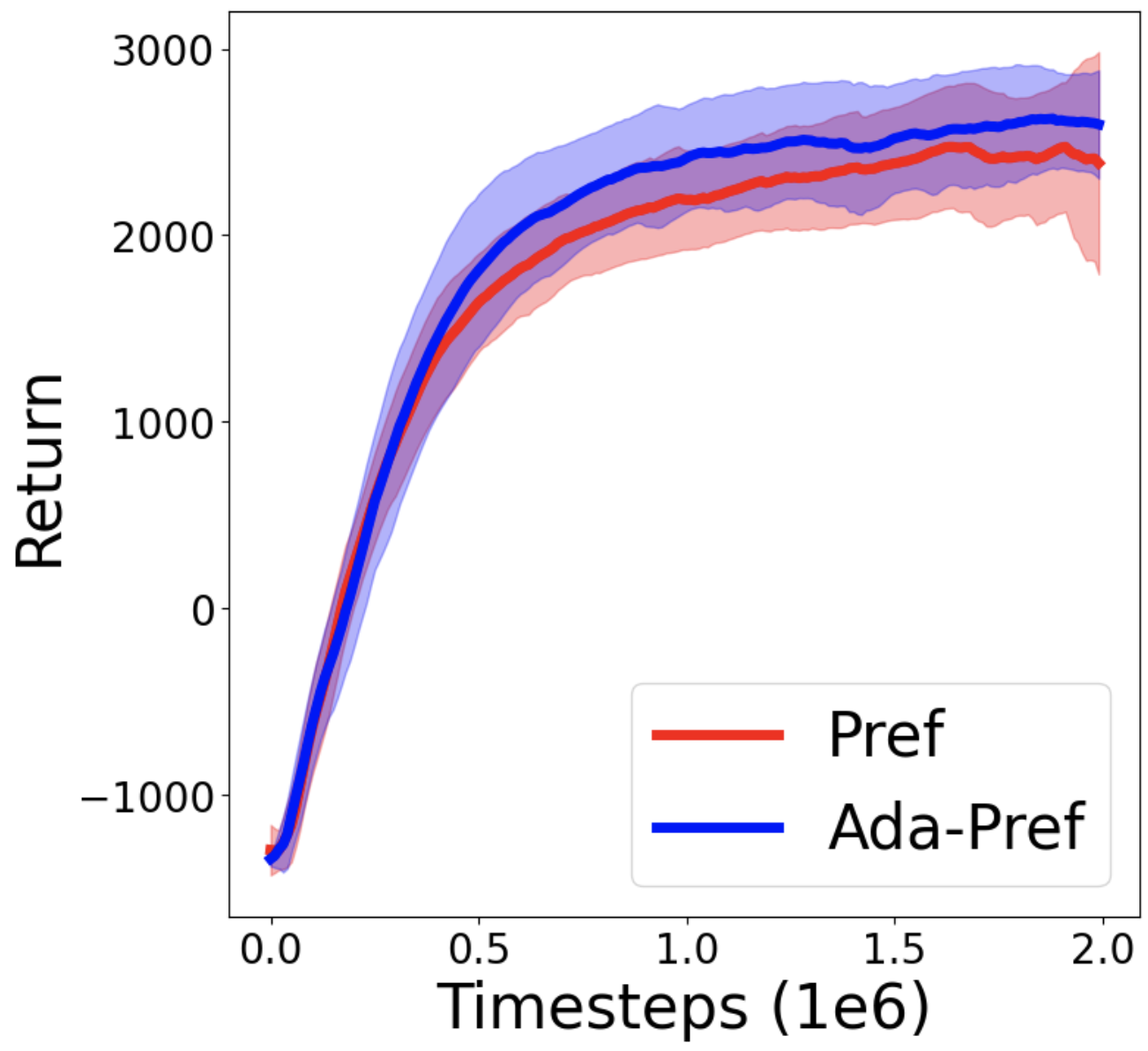}
\end{minipage}
\begin{minipage}[t]{0.26\linewidth}
\includegraphics[width=0.97\linewidth]
{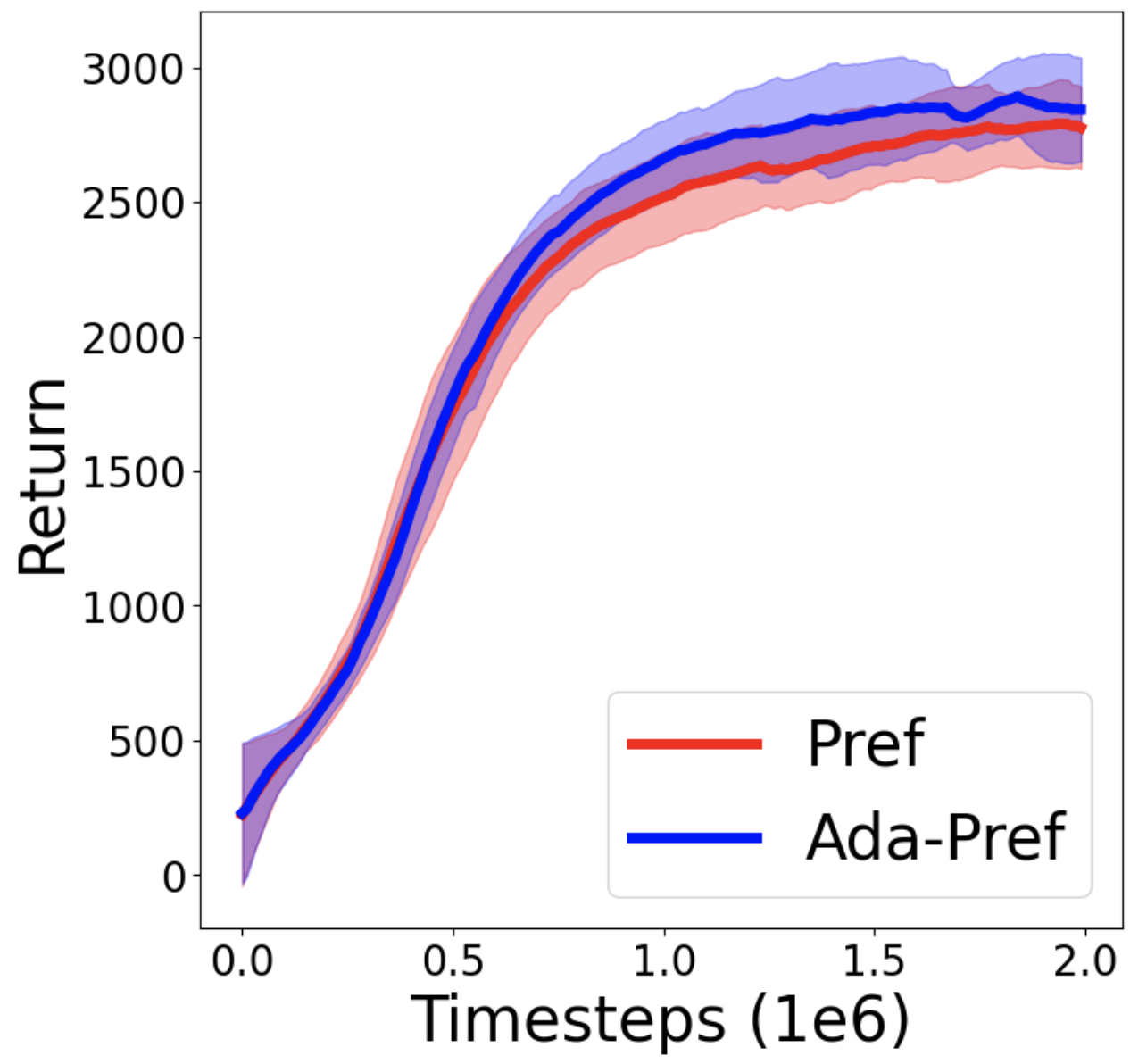}
\end{minipage}
\begin{minipage}[t]{0.26\linewidth}
\includegraphics[width=0.97\linewidth]
{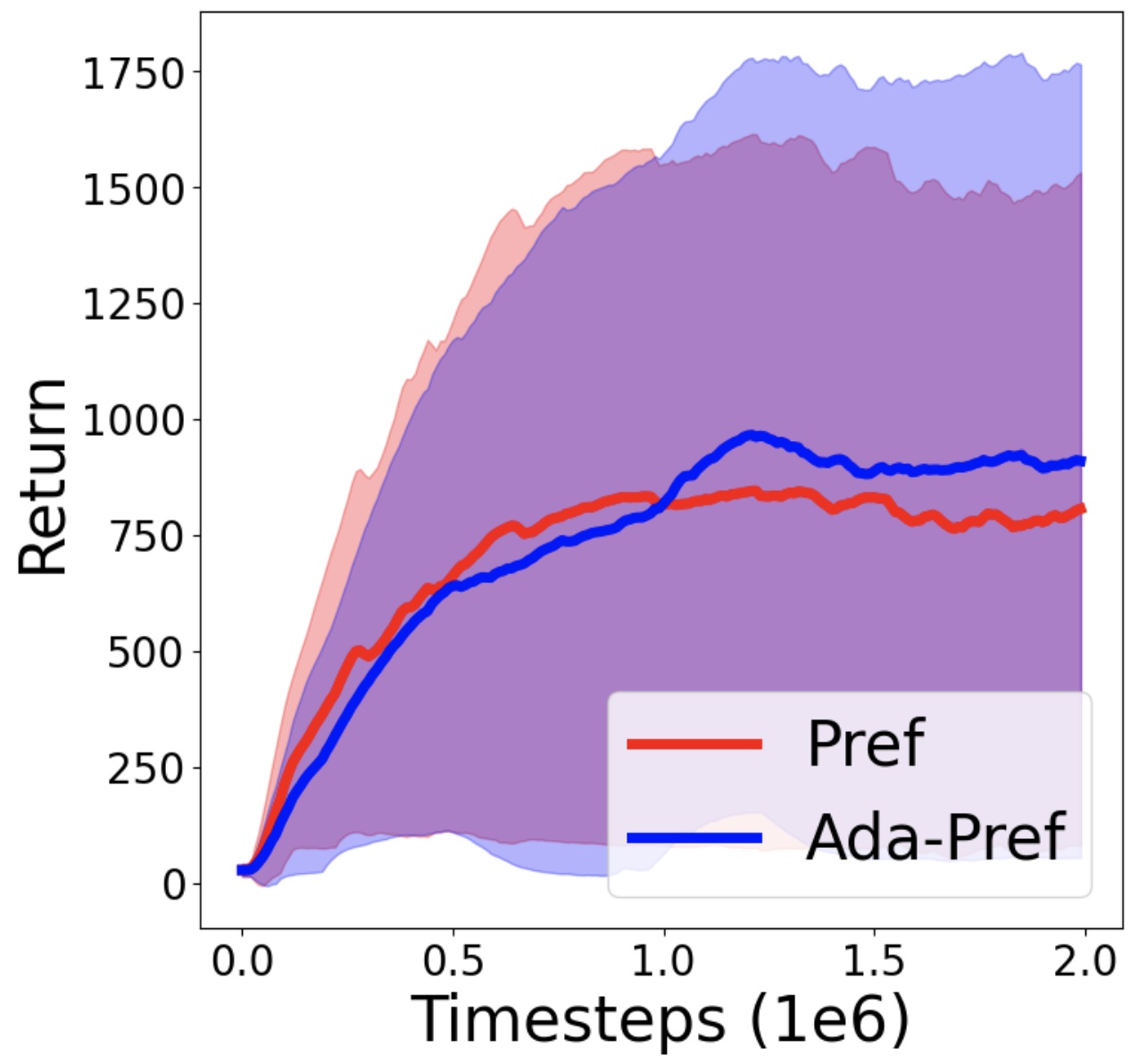}
\end{minipage}
\\
%\vspace{-0.1in}
\subfigure[HalfCheetah]{
\centering\includegraphics[width=0.25\linewidth]
{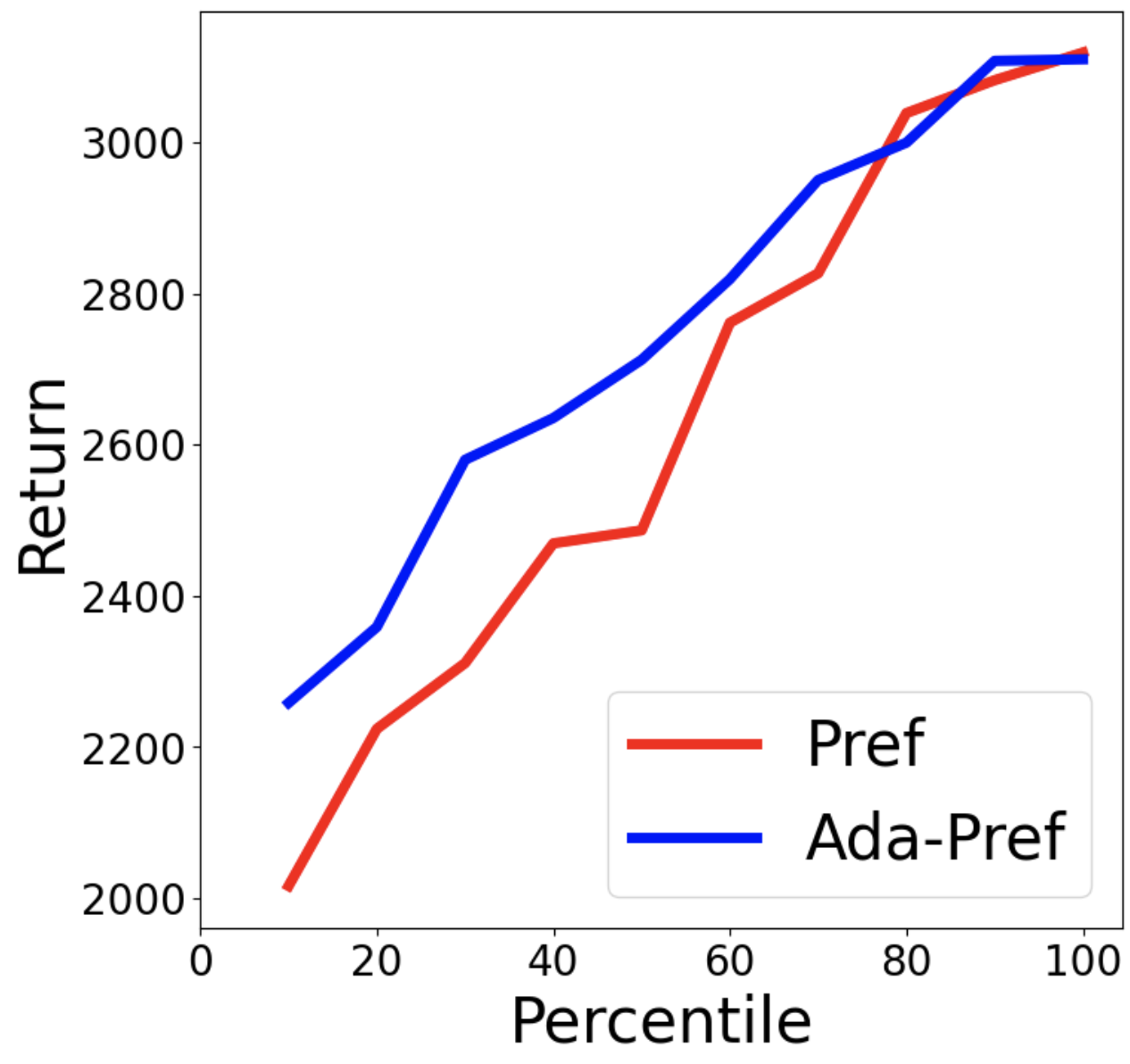}
}
\subfigure[Ant]{
\centering\includegraphics[width=0.25\linewidth]
{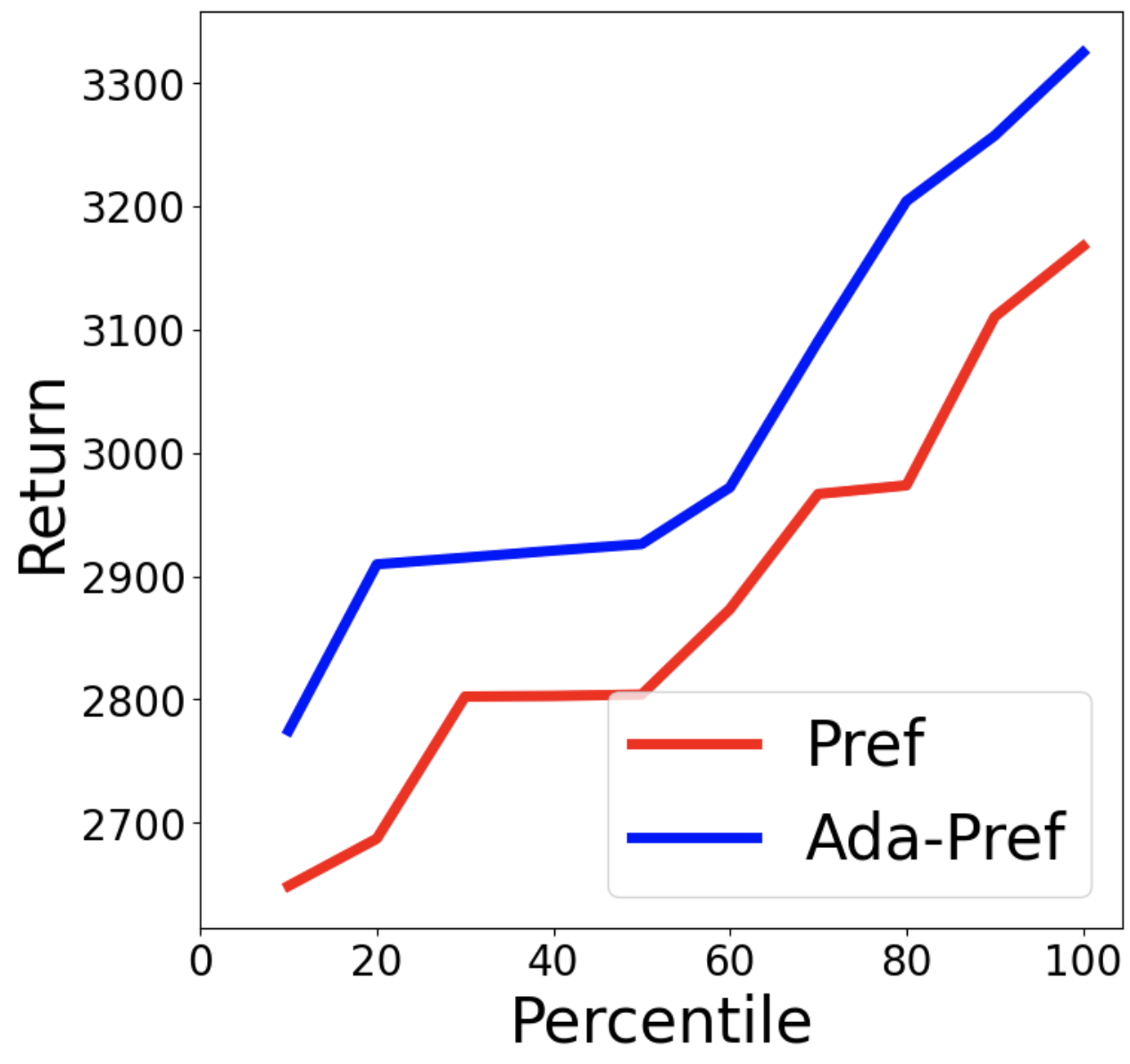}
}
\subfigure[Hopper]{
\centering\includegraphics[width=0.25\linewidth]
{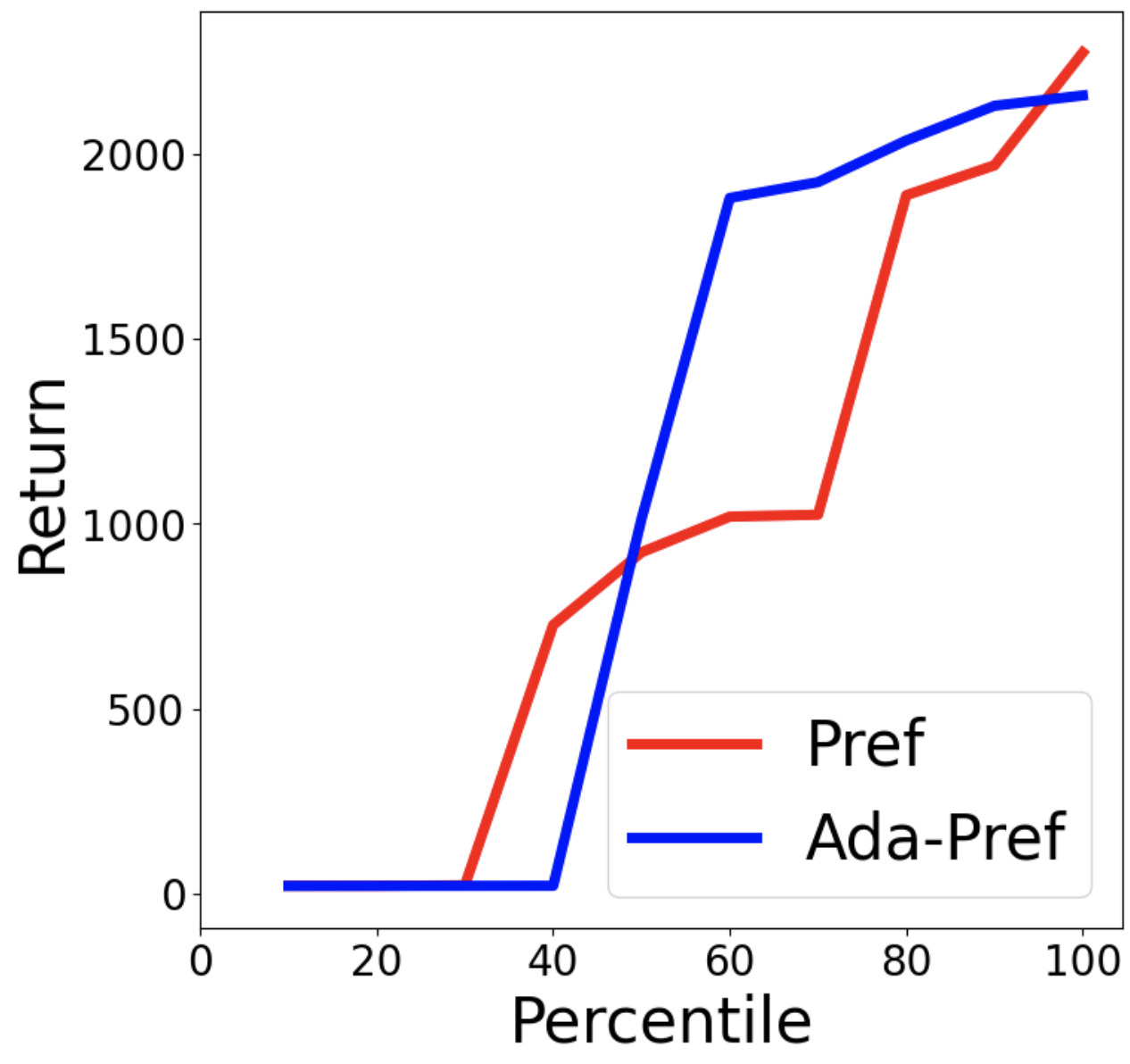}
}
%%\vspace{-0.1in}
\caption{Learning curve plots (top) and percentile plots (bottom) for Pref and Ada-Pref. For the learning curve plots, returns at each timestep are averaged across 10 different seeds, then smoothed over timesteps using an exponential moving average (EMA) with a smoothing factor of $\alpha=0.1$. For the percentile plots, returns from 10 different seeds are sorted in ascending order.
}
\label{fig:pybullet_exp2}
\end{figure}

We present the results for three PyBullet tasks, using a segment size of 25. Table \ref{tab:pybullet3} and Figure \ref{fig:pybullet_exp2} show the performance of Pref and Ada-Pref on the PyBullet tasks, based on the first hyperparameter tuning criterion. Table \ref{tab:pybullet4} displays the results for Pref and Ada-Pref according to the second hyperparameter tuning criterion. These results reconfirm the effectiveness of our adaptive preference loss.

\begin{table}[htb!]
\centering
\resizebox{.99\linewidth}{!}{%
\begin{minipage}{.48\linewidth}
\caption{Table for the highest return of the best policy and the average preference prediction accuracy of the corresponding reward function.}
\label{tab:pybullet3}
\centering
\begin{tabular}{clcc}
\toprule
\multirow{2}{*}{\bf Task}  & \multirow{2}{*}{\bf Method}  & \multirow{2}{*}{\bf Return}  & {\bf Preference}   \\ 
~ & ~ & ~ & {\bf Accuracy (\%)}  \\\midrule
\multicolumn{1}{c|}{\multirow{2}{*}{HalfCheetah}}   & \multicolumn{1}{l|}{\baseline}         & 2575.69           & 90.82                \\
\multicolumn{1}{c|}{}                               & \multicolumn{1}{l|}{\ourmethod}    & \textbf{2689.9}  & 90.35       \\\midrule
\multicolumn{1}{c|}{\multirow{2}{*}{Ant}}           & \multicolumn{1}{l|}{\baseline}         & 2832.87           & 84.88                \\
\multicolumn{1}{c|}{}                               & \multicolumn{1}{l|}{\ourmethod}    & \textbf{2960.47}  & 84.09      \\\midrule
\multicolumn{1}{c|}{\multirow{2}{*}{Hopper}}        & \multicolumn{1}{l|}{\baseline}         & 883.49           & 85.0                \\
\multicolumn{1}{c|}{}                               & \multicolumn{1}{l|}{\ourmethod}    & \textbf{1025.74}  & 85.15        \\
\bottomrule
\end{tabular}
\end{minipage}
\hspace{0.5cm}
\begin{minipage}{.48\linewidth}
\centering
\caption{Table for the average preference prediction accuracy of the best reward function and the highest return of the corresponding policy.}\label{tab:pybullet4}
\centering
\begin{tabular}{clcc}
\toprule
\multirow{2}{*}{\bf Task}  & \multirow{2}{*}{\bf Method}  & \multirow{2}{*}{\bf Return}  & {\bf Preference}   \\ 
~ & ~ & ~ & {\bf Accuracy (\%)}  \\\midrule
\multicolumn{1}{c|}{\multirow{2}{*}{HalfCheetah}}   & \multicolumn{1}{l|}{\baseline}         & 2564.49           & 91.38                \\
\multicolumn{1}{c|}{}                               & \multicolumn{1}{l|}{\ourmethod}    & \textbf{2609.03}  & 90.79       \\\midrule
\multicolumn{1}{c|}{\multirow{2}{*}{Ant}}           & \multicolumn{1}{l|}{\baseline}         & 2738.4           & 86.21                \\
\multicolumn{1}{c|}{}                               & \multicolumn{1}{l|}{\ourmethod}    & \textbf{2917.22}  & 85.15     \\\midrule
\multicolumn{1}{c|}{\multirow{2}{*}{Hopper}}        & \multicolumn{1}{l|}{\baseline}         & 796.52           & 85.79                \\
\multicolumn{1}{c|}{}                               & \multicolumn{1}{l|}{\ourmethod}    & \textbf{1025.74}  & 85.15        \\
\bottomrule
\end{tabular}
\end{minipage}%
}
\end{table}

\end{document}